\documentclass{article}

\usepackage[final]{corl_2024} 

\usepackage[numbers]{natbib}
\usepackage{multicol}
\usepackage{graphicx}
\usepackage{graphics}
\usepackage{wrapfig}
\usepackage{amsmath}
\usepackage{amssymb}
\usepackage{amsthm}
\usepackage{tabularx}
\theoremstyle{definition}

\usepackage[linesnumbered,ruled,vlined]{algorithm2e}
\usepackage{multirow}
\usepackage{color}
\definecolor{lightgreen}{rgb}{0.0,0.7,0.0} 

\usepackage{array}
\usepackage{makecell}

\newcommand{\um}[1]{{#1}} 

\newcommand{\Ours}{GFC\xspace}

\title{Generative Factor Chaining: Coordinated Manipulation with Diffusion-based Factor Graph}

\author{
  Utkarsh A. Mishra, Yongxin Chen, Danfei Xu\\
  Georgia Institute of Technology \\
  \texttt{\{umishra31, yongchen, danfei\}@gatech.edu} \\
}

\begin{document}
\maketitle



\begin{abstract}
    Learning to plan for multi-step, multi-manipulator tasks is notoriously difficult because of the large search space and the complex constraint satisfaction problems. We present Generative Factor Chaining~(GFC), a composable generative model for planning. GFC represents a planning problem as a spatial-temporal factor graph, where nodes represent objects and robots in the scene, spatial factors capture the distributions of valid relationships among nodes, and temporal factors represent the distributions of skill transitions. Each factor is implemented as a modular diffusion model, which are composed during inference to generate feasible long-horizon plans through bi-directional message passing. We show that GFC can solve complex bimanual manipulation tasks and exhibits strong generalization to unseen planning tasks with novel combinations of objects and constraints. More details can be found at: \href{https://generative-fc.github.io/}{generative-fc.github.io}
\end{abstract}




\section{Introduction}



\begin{figure}[h]
    \centering
    \includegraphics[width=0.95\linewidth]{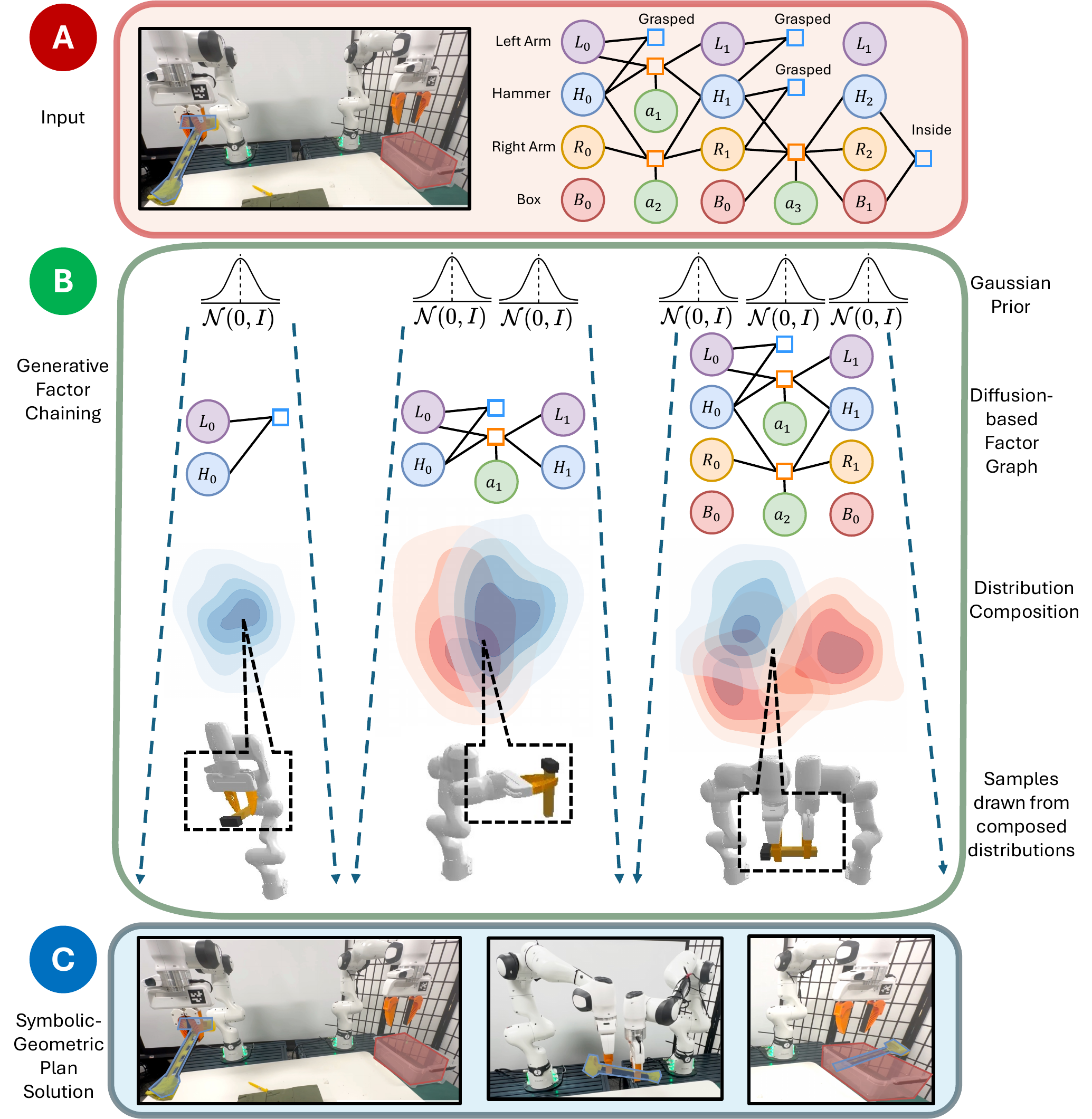}
    \caption{\textbf{Factor graph for a multi-arm coordination task.} Our factor graph-based planning formulation solves for a sequence of spatial factor graphs from the initial state to a goal factor by chaining them using temporal skill factors. The figure illustrates the temporal evolution of a factor graph by executing single or multiple skills sequentially or in-parallel to handover a hammer, pick up a nail, and coordinate both arms to strike the nail. \textbf{Task:} The task objective is to place the hammer inside the box. However, since the left arm cannot reach the box, the hammer is handed over to the right arm such that the right arm can complete the task. \textbf{(a)~Inputs:} The initial scene and a symbolically feasible spatial-temporal factor graph plan to complete the goal objective. \textbf{(b)~GFC:} We formulate all factors as distributions of the nodes connected to them. GFC represents spatial factors as classifiers and temporal factors as diffusion models. We leverage compositionality of diffusion models to compose spatial-temporal distributions and find the joint distribution of the complete plan directly at inference. Finally, samples drawn from such a joint distribution are symbolically and geometrically feasible solutions of the whole plan. \textbf{(c)~Output:} A sequence of skill choices and optimizer continuous parameters executed on robots with parameterized skill controllers.}
    \label{fig:method_figure}
\end{figure}

Solving real-world sequential manipulation tasks requires reasoning about sequential dependencies among manipulation steps. For example, a robot needs to grip the center or the tail of a hammer, instead of its head, in order to subsequently hammer a nail. The complexity of planning problems increases when multiple manipulators are involved, where spatial coordination constraints among manipulators need to be satisfied. In the example shown in~\autoref{fig:method_figure}, the robot has to reason about the optimal pose to grasp the hammer with the left arm, such that the right arm can coordinate to re-grasp. Subsequently, the two arms must coordinate to hammer the nail. 
While classical Task and Motion Planning~(TAMP) methods have shown to be effective at solving such problems by hierarchical decomposition~\cite{garrett2021integrated}, they require accurate system state and kinodynamic model. Further, searching in such a large solution space to satisfy numerous constraints poses a severe scalability challenge. In this work, we aim to develop a learning-based planning framework to tackle complex manipulation tasks with both sequential and spatial coordination constraints.


To solve complex sequential manipulation problems, prior learning-to-plan methods have largely adopted the options framework and modeled the preconditions and effect of the options or primitive skills~\cite{bacon2017option,driess2021learning,xu2021deep,agia2022taps,mishra2023generative,liang2022search}. Key to their successes are skill chaining functions that determine whether executing a skill can satisfy the precondition of the next skill in the plan, and eventually the success condition of the overall task. However, the use of vectorized states and the assumption of a linear chain of sequential dependencies limits the expressiveness of these methods. Consider a task where a robot fetches two items from a box. Intuitively, the skills for fetching one object should not influence the other. However, due to vectorized states and the linear dependency assumption, the skill-chaining methods are forced to model such sequential dependencies. Similarly, a skill intended to satisfy a future skill's condition will be forced to influence the steps in between. Finally, the skill chain representation forbids these methods from effectively modeling multiple-arm manipulation tasks, where concurrent skills must be planned to jointly satisfy a constraint.

To move beyond the linear chain and model complex coordinated manipulation, we introduce Generative Factor Chaining (GFC), a learning-to-plan framework built on flexible composable generative models. \um{For a given symbolically feasible plan graph}, GFC adopts a spatial-temporal factor graph~\cite{dellaert2021factor} representation, where nodes are objects and robot states, and spatial factors represent the relationship constraints between these nodes. Skills are temporal factors that connect these state-factor graphs via transition distributions. 
A single skill factor can simultaneously connect to multiple object and robot nodes, allowing for natural representation of complex multi-object interactions and steps that necessitate coordination between multiple manipulators.
During inference, this factor graph can be treated as a probabilistic graphical model, where the learned skill factor and spatial constraint factor distributions are composed to form a joint distribution of complete plans. Through 13 long-horizon manipulation tasks in simulation and the real world, we show that GFC can solve complex bimanual manipulation tasks and exhibits strong generalization to unseen planning tasks with novel combinations of objects and constraints.

\section{Related Work}
\label{sec:related}


\textbf{Task and Motion Planning (TAMP).} 
TAMP frameworks decompose a complex planning problem into constraint satisfaction problems at task and motion levels~\cite{sutton1999between,bacon2017option,shah2021value, nachum2018data, masson2016reinforcement}. 
Notably, Garret et al.~\cite{garrett2021integrated} drew connections between TAMP and factor graphs~\cite{dellaert2021factor}, representing constraints as factors and objects/robots as nodes. This formalism naturally allows reusing per-constraint solvers across tasks. 
However, classical TAMP approaches often rely on accurate perception and system dynamics, limiting their practical applications and scalability. We instead opt for a learning approach, while our compositional factor graph representation remains heavily inspired by the classical TAMP paradigm.

\textbf{Generative models for planning.} Modern generative models have been applied to offline imitation~\cite{yu2023scaling, kapelyukh2023dall, pearce2023imitating, Chi2023DiffusionPV,mishra2023reorientdiff,Du2023LearningUP, reuss2023goal, reuss2023multimodal} and reinforcement learning~\cite{janner2022diffuser, ajay2022conditional}. In addition to modeling complex state and action distributions, generative models have also been shown to encourage compositional generalization~\cite{Zhang2023DiffCollagePG,mishra2023generative,du2020compositional} by combining data across tasks~\cite{ajay2022conditional,janner2022diffuser}. Most relevant to us are Generative Skill Chaining~(GSC)~\cite{mishra2023generative} and Diffusion-CCSP~\cite{yang2023compositional}, both designed to achieve systematic compositional generalization. GSC composes skill chains through a guided diffusion process. However, similar to other skill-chaining methods~\cite{xu2021deep,agia2022taps}, GSC cannot model non-linear dependencies such as parallel skills and independence among skills. 
Diffusion-CCSP trains diffusion models to generate object configurations to satisfy  spatial constraints, while relying on external solvers to plan the manipulation sequence. 
Our method is a unified framework to solve the combined problem: it generates skill plans to satisfy both spatial and temporal constraints represented in a factor graph.

\textbf{Learning for coordinated manipulation.} Coordinating two or more arms for manipulation presents numerous planning challenges~\cite{chen2022cooperative,nagele2020legobot,ureche2018constraints}, including the combinatorial search space complex constraints for coordinated motion. Recent works have utilized learning-based frameworks~\cite{amadio2019exploiting,chitnis2020efficient,ha2020learning,grannen2023stabilize,tung2021learning} in both Reinforcement Learning~\cite{amadio2019exploiting,ha2020learning} and offline Imitation Learning~\cite{tung2021learning,grannen2023stabilize}. However, most existing works have focused on learning task-specific policies~\cite{amadio2019exploiting,grannen2023stabilize} or require multi-arm demonstration data collected through a specialized teleoperation device~\cite{tung2021learning}. In contrast, our factor graph-based representation enables solving multi-arm tasks by composing multiple single-arm skills through inference-time optimization. 


\section{Background}
\label{sec:background}

\textbf{Diffusion Models.} A core component of our method is based on distributions learned using diffusion models. A diffusion model learns an unknown distribution~$p(\textbf{x}^{(0)})$ from its samples by approximating the score function~$\nabla \log p$. It consists of two processes: a \textit{forward diffusion or noising} process that progressively injects noise and a \textit{reverse diffusion or denoising} process that iteratively removes noise to recover clean data. The forward process simply adds Gaussian noise $\epsilon$ to clean data as $\textbf{x}^{(t)}=\textbf{x}^{(0)}+\sigma_t \epsilon$ for a monotonically increasing $\sigma_t$. The reverse process relies on the score function $\nabla_\textbf{x}\log p_t(\textbf{x}^{(t)})$ where $p_t$ is the distribution of noised data $\textbf{x}^{(t)}$.
In practice, 
the unknown score function is estimated using a neural network $\epsilon_\phi(\textbf{x}^{(t)}, t)$ by minimizing the denoising score matching~\cite{song2020score} objective 
$
\mathbb{E}_{t,\epsilon, \textbf{x}^{(0)}}[ \lambda(t) \| \epsilon -  \epsilon_\phi(\textbf{x}^{(t)}, t) \|^2 ]
$
where $\lambda(t)$ is a time-dependent weight. Several recent works have explored the advantages of diffusion models like scalability~\cite{ho2020denoising, song2020denoising, zhang2022gddim, zhang2022fast} and the ability to learn multi-modal distributions~\cite{ho2022classifier, dhariwal2021diffusion, lugmayr2022repaint, ajay2022conditional}. We are particularly interested in the compositional ability~\cite{Zhang2023DiffCollagePG, yu2023scaling, du2020compositional, yang2023compositional, mishra2023generative} of these models for the proposed method.

\textbf{Problem setup.} We assume access to a library of parameterized skills~\cite{kaelbling2017learning} $\pi\sim\Pi$ such as primitive actions like \texttt{Pick} and \texttt{Place}. Each skill $\pi$ requires a pre-condition to be fulfilled and is parameterized by a continuous parameter $a \in A_\pi$ governing the desired motion while executing the skill in a state $s$. 
For a given symbolically feasible task plan from a starting state $s_0$ to reach a specified goal condition $s_{goal}$, generated by a task planner or given by an oracle, the problem is to obtain the sequence of continuous parameters to make the plan geometrically feasible. For example, given a nail at a target location and a hammer on a table, the symbolic plan is to \texttt{Pick} the hammer and \texttt{Reach} the nail. A geometrically-feasible plan requires suitable \texttt{Pick} and \texttt{Reach} parameters such that the hammer's head can strike the nail. 

\textbf{Learning for skill chaining.} Existing works along this direction model the planning problem as a ``chaining'' problem: They first model the pre-conditions and effect state distributions for every skill~$\pi\sim\Pi$ from the available data and a symbolic {\em plan skeleton}~$\Phi_K = \{\pi_1, \pi_2, ..., \pi_K\}$ consisting of $K$-skills is constructed. With this model, they search for the given skill sequence~(plan) such that each skill satisfies the pre-conditions of the next skill in the plan. STAP~\cite{agia2022taps} used learned priors to perform data-driven optimization with the cross-entropy maximization method. In GSC~\cite{mishra2023generative}, the policy and transition model is formulated as a diffusion model based distribution~$p_\pi(s, a_\pi, s')$ which allows for flexible chaining. While the forward chain ensures dynamics consistency in the plan, backward chain ensures that the goal is reachable from the intermediate states. For a forward rollout trajectory~$\tau=\{s_0, a_{\pi_1}, s_1, a_{\pi_2}, s_{goal}\}$ associated with skeleton $\Phi_2 = \{\pi_1, \pi_2\}$, 
the resulting forward-backward combination based on GSC~\cite{mishra2023generative} can be represented as 
\begin{equation}
\label{eq:pgm-gsc}
p_\tau(\tau|s_0, s_{goal}) \propto \frac{p_{\pi_1}(s_0, a_{\pi_1}, s_1)p_{\pi_2}(s_1, a_{\pi_2}, s_{goal})}{\sqrt{p_{\pi_1}(s_1)p_{\pi_2}(s_1)}}
\end{equation}

\section{Method}
We aim to solve unseen long-horizon planning problems by exploiting the inter-dependencies between the objects important for the task at hand in the scene. Our method adopts factor graphs to represent states and realize their temporal evolution by the application of skills. While previous works have considered {\em vectorized state} representations making it difficult to decouple spatial-independence, we focus on {\em factorized state} representations such that the state of the environment is entirely modular, containing information about all the objects in the scenario and the task-specific constraints between them. We use a spatial-temporal factor graph~\cite{dellaert2021factor} that is transformed into a probabilistic graphical model by representing temporal factors as skill-level transition distributions and spatial factors as constraint-satisfaction distributions. A composition of all the factors jointly represents sequential and coordinated manipulation plans directly at inference and can be solved by sampling optimal node variables using reverse diffusion sampling.

\subsection{Representing States, Skills, and Plans in Factor Graphs}
\textbf{States as factor graphs.}  We define a factor graph $\{\mathcal{V}, \mathcal{F}\}$ of a state $s$ consisting of the decision variable $\mathcal{V}$ and factor $\mathcal{F}$ nodes. Every robot and object is represented as a decision variable node $v \in \mathcal{V}$ containing their respective state. Factors $f \in \mathcal{F}$ between nodes in a given state are \emph{spatial constraints}. For example, a \texttt{Grasped} spatial factor specifies admissible rigid transforms between a gripper and an object.  
When we construct a probabilistic graphical model from the representation described above, an intuitive way of calculating the distribution of a state,~$p(s)$, is the composition of all the factor distributions. Mathematically: 
\begin{equation}
\label{eq:state_dist}
    p(s) \propto \prod_{f \in \mathcal{F}} p_f(\mathcal{S}_f) \quad \text{ where } s \equiv \bigcup_{f \in \mathcal{F}} \mathcal{S}_f
\end{equation}
where $p_f(\mathcal{S}_f)$ represents the joint factor potential of nodes $v \in \mathcal{S}_f \subseteq \mathcal{V}$, i.e. all nodes involved in a factor \footnote{i.e. a factor $f$ is included \textit{iff} there is an edge between $f$ and some $v \in \mathcal{V}$ which also implies $v \in \mathcal{S}_f \subseteq \mathcal{V}$.} and $s$ is the joint distribution of all such nodes. 
This indicates that the joint distribution of all the nodes must satisfy each of the factors, also explored by Diffusion-CCSP~\cite{yang2023compositional}.

\textbf{Skills as temporal factors.} To represent transitions between states, we adapt parameterized skills~\cite{kaelbling2017learning} for a factor graph formulation. We define the preconditions of a skill as a set of nodes and factors, thus considering a skill feasible \emph{iff} the precondition factors are satisfied. For example, for state $s_0$ illustrated  in~\autoref{fig:method_figure}, the nodes of a factor graph are \{$L_0, H_0, R_0, B_0$\} and the factors existing in this scene are \{\texttt{Grasped($L_0, H_0$)=True}\}. Now, since this factor is a precondition of the skill \texttt{Move}$(L_0, H_0)$ that moves the hammer in hand to align with the box, it must be satisfied for the skill to be feasible. The effect of executing a skill creates a new factor graph $s'$ by changing the state of the nodes involved and, optionally, adding or removing their factors.  This results in a \emph{temporal factor} between the transitioned nodes of $s$ and $s'$ with the continuous action parameter of the skill $a_{\pi}$. 
The skill definitions can be extracted from standard PDDL symbolic skill operator with minor adaptations, following the duality of factor graphs and plan skeletons~\cite{garrett2021integrated}. Eventually, we solve an optimization problem: satisfying the \texttt{Aligned}, \texttt{Grasped}, and the transition dynamics constraints by finding the correct \texttt{Move} parameters $a_{\pi_1}$. Each skill in a plan introduces additional nodes and factors to the factor graph, with added complexity for optimization. 

Mathematically, we can use the distribution $p(s)$ as established in~\autoref{eq:state_dist} with all the spatial factors, and represent the temporal skill factor distribution of $k^{th}$-skill $\pi_k$ as the joint distribution: 
$
p_{\pi_k}(s,a,s') \equiv p_{\pi_k}(S_{\pi_k}, a, S'_{\pi_k}),\;\; S_{\pi_k} \subseteq \mathcal{V}^{\pi_k}_{pre}
$
which is executable \textit{iff} the skill's pre-condition $s^{\pi_k}_{pre}~\equiv~\{\mathcal{V}^{\pi_k}_{pre}, \mathcal{F}^{\pi_k}_{pre}\}$ is satisfied by the current state i.e. $\mathcal{V}^{\pi_k}_{pre} \subseteq \mathcal{V}$ and $\mathcal{F}^{\pi_k}_{pre} \subseteq \mathcal{F}$.
Once executed, it leads to the transitioned state $S'_{\pi_k}$.
Based on the above formulation of a short-horizon transition distribution, we extend to construct a plan-level distribution as already established by GSC~\cite{mishra2023generative} and shown in~\autoref{eq:pgm-gsc}. 
We leverage the modularity of factored states by replacing states $s$ with a set of decision variables $S_{\pi_k}$ in the interest of skill $\pi_k$. This allows us to chain multiple skills in series and parallel. In such a scenario, the denominator term exists only for certain decision nodes \textit{iff} they are common in two consecutive skills. We can indeed rewrite~\autoref{eq:pgm-gsc} as:
\begin{equation}
\label{eq:skill_dist}
    p(\tau) \propto \frac{\prod_{\pi_k \in \Phi} p_{\pi_k}(v_k \in \mathcal{V}^{\pi_k}_{pre}, a_k, v'_{k} \in \mathcal{V}^{\pi_k}_{effect})}{\sqrt{\prod_{v_i \in \mathcal{V}_i} p_{\pi_{i-}}(v_i) p_{\pi_{i+}}(v_i)}}
\end{equation}
if we consider that some set of intermediate nodes $\mathcal{V}_i$ are connected by two sequential skills $\pi_{i-}$ and $\pi_{i+}$.


\begin{figure*}[t]
    \centering
    \includegraphics[width=0.9\linewidth, trim={0, 0.5cm, 0cm, 0cm}, clip]{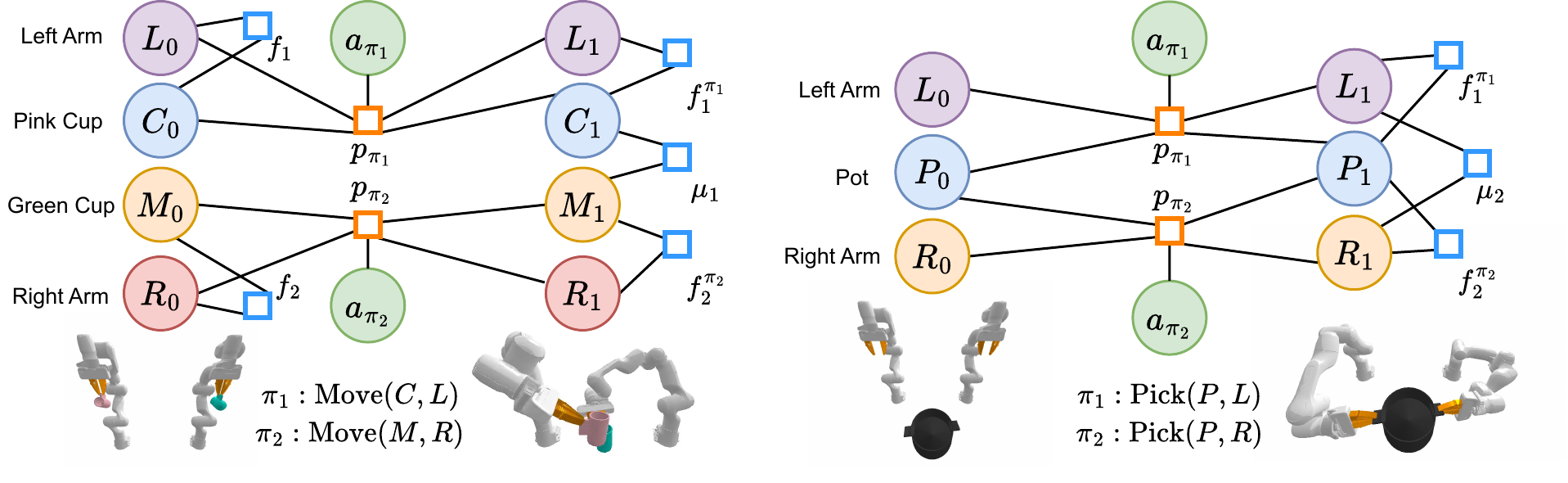}
    \caption{(Left) \textbf{Parallel independent chaining} The figure shows the execution of two skills~($\pi_1$ and $\pi_2$) in-parallel on two independent sets of nodes (L, C and R, M) to modify their existing factors~(\texttt{Grasped}). The two independent executions can be connected via external factors~$\mu_1$~(\texttt{FixedTransform}) introducing spatial dependencies between nodes C and M. (Right)~\textbf{Parallel dependent chaining} The figure shows overlapping nodes of interest while parallel execution of two skills. The pot is to be picked by using both arms simultaneously. The effect of this is resulting factors~(\texttt{Grasped}) between (L, P and R, P) and external factor $\mu_2$~(\texttt{FixedTransform}) between L and R. Overlapping nodes satisfy both skill's temporal effects.}
    \label{fig:parallel_chaining_example}
\end{figure*}

\textbf{Representing coordination.} A key advantage of the factor graph representation is the ability to model multi-arm coordination tasks by connecting the temporal chains of each arm using spatial constraints. Such tasks often require skills to be simultaneously executed on each arm to operate on different or the same objects.
We consider two cases for parallel skill execution, where multiple robots are operating on: (1) independent objects and (2) the same object, leading to independent and dependent temporal chains respectively. 
With our factorized state representation, we can independently control the execution of individual skills correlated with the nodes of interest and calculate the cumulative effect by applying the union of the effects of all the skills to the current factor graph. 
We consider a scenario shown in~\autoref{fig:parallel_chaining_example}~(Left). The left and right gripper arm $L_0$ are holding the pink $C_0$ and green $M_0$ cup~(\{\texttt{Grasped($L_0, C_0$)=True}\} and \{\texttt{Grasped($R_0, M_0$)=True}\}) respectively. While both the grippers can independently execute the skill \texttt{Move} to modify separate factors~($f^{\pi_1}_1$ and $f^{\pi_2}_2$), one can add a constrained relationship factor~($\mu_1$) between the two mugs representing a set of transforms that satisfy the precondition of \texttt{Pour}. Such an ability to augment constraints flexibly allows zero-shot coordination planning for unseen tasks at test time even with parallel skill executions on the same object as shown in~\autoref{fig:parallel_chaining_example}~(Right).

\subsection{Generative Factor Chaining} 
\label{ssec:gfc}

Now we have a formulation to construct a symbolic spatial-temporal factor graph plan for a task and chain them using spatial factor and temporal skill factors sequentially or in parallel. To make this plan geometrically feasible, we must find the optimal node variable values. While classical solvers require modeling the transition dynamics of complex manipulation tasks, sampling-driven optimization with learned models provides less flexibility and modularity~\cite{mishra2023generative}. In this work, we leverage the expressive generative model to capture the transition dynamics and exploit the compositionality of diffusion models. \um{Given a symbolically feasible factor graph plan,} our method, termed Generative Factor Chaining (\Ours), can flexibly compose spatial-temporal factor distributions to sample optimal node variable values for the complete plan. 

\textbf{Probabilistic model for trajectory plan as spatial-temporal factor graphs}. Now, we again consider the spatial graph for representing the state, where the probability of finding a state $s$ is the joint distribution of all the nodes in the factor graph. We will now integrate the spatial factors with the temporal factors considering the compensation term introduced in \autoref{eq:state_dist} and \autoref{eq:skill_dist} along with the constraint factors across the chain $\mu \in \mathcal{M}$ as:
\begin{equation}
\label{eq:final_gfc}
    p(\tau) \propto \frac{\prod_{\pi_k \in \Phi} p_{\pi_k}(v_k \in \mathcal{V}^{\pi_k}_{pre}, a_k, v_{k+1} \in \mathcal{V}^{\pi_k}_{effect}) \prod_{k=0}^K\prod_{f \in \mathcal{F}_k} p_f(\mathcal{S}_f)}{\sqrt{\prod_{v_i \in \mathcal{V}_i} p_{\pi_{i-}}(v_i) p_{\pi_{i+}}(v_i)}} \Pi_{\mathcal{M}} f_\mu(S_\mu)
\end{equation}
This completes the joint distribution of all the nodes in the spatial-temporal factor graph plan considering the temporal factors for all skills with their pre-condition and effect nodes, all spatial factors for all states in the plan, and all intermediate nodes in the temporal chain. We show our implementation of this formulation in \autoref{algo:gfc}.

For the sake of simplicity, we will formulate the probabilistic model for the two chains shown in~\autoref{fig:parallel_chaining_example} by following the forward-backward analysis introduced by GSC and discussed in~\autoref{sec:background}. We can write the top chain as:
\begin{equation}
\label{eq:pgm-ex1}
p_{\pi_1}(L_0, C_0, a_{\pi_1}, L_1, C_1)p_{\pi_2}(R_0, M_0, a_{\pi_2}, R_1, M_1) p_{\mu_1}(C_1, M_1)
\end{equation}
showing the independence of factors. Similarly, the bottom chain can be constructed based on~\autoref{eq:final_gfc} as:
\begin{equation}
\label{eq:pgm-ex2}
\frac{p_{\pi_1}(L_0, P_0, a_{\pi_1}, L_1, P_1)p_{\pi_2}(R_0, P_0, a_{\pi_2}, R_1, P_1)}{\sqrt{p_{\pi_1}(P_1)p_{\pi_2}(P_1)}} p_{\mu_2}(L_1, R_1)
\end{equation}
where the factors are dependent on each other. It is worth noting that the augmented constraint factors $p_{\mu}$ work as a weighing function and can be more precisely represented by $p_{\mu}(S_\mu) \equiv p_{\mu}(y=1|S_\mu)$ for some constraint-satisfaction index $y$. 

We align towards diffusion model-based learned distributions to represent the probabilities in the formulated probabilistic graphical model. We transform the probabilities into their respective score functions $\epsilon(\textbf{x}^{(t)}, t)$ for a particular reverse diffusion sampling step $t$ and train it using score matching loss. Hence, for sampling a scene-graph for~\autoref{eq:final_gfc}, we have
\begin{gather*}
\label{eq:gfc_prob_model}
\epsilon(\tau^{(t)}, t) = \sum_{\pi_k \in \Phi} \epsilon_{\pi_k}(v^{(t)}_k \in \mathcal{V}^{\pi_k}_{pre}, a^{(t)}_k, v^{(t)}_{k+1} \in \mathcal{V}^{\pi_k}_{effect}, t) + \sum_{k=0}^K\sum_{f \in \mathcal{F}_k} \epsilon_f(\mathcal{S}^{(t)}_f, t) \\ 
- \frac{1}{2} \sum_{v_i \in \mathcal{V}_i} \Big[\epsilon_{\pi_{i-}}(v^{(t)}_i, t) \epsilon_{\pi_{i+}}(v^{(t)}_i, t)\Big] +\sum_{\mathcal{M}} \epsilon_{f_\mu}(S^{(t)}_\mu, t)
\end{gather*}
Following this, we can show for the dependent factor chain in~\autoref{eq:pgm-ex2} as:
\begin{gather*}
    \epsilon(L^{(t)}_0, P^{(t)}_0, R^{(t)}_0, L^{(t)}_1, P^{(t)}_1, R^{(t)}_1, t) = 
\epsilon_{\pi_1}(L^{(t)}_0, P^{(t)}_0, a^{(t)}_{\pi_1}, L^{(t)}_1, P^{(t)}_1, t) + \\
\epsilon_{\pi_2}(R^{(t)}_0, P^{(t)}_0, a^{(t)}_{\pi_2} R^{(t)}_1, P^{(t)}_1, t) - \frac{1}{2} \epsilon_{\pi_1}(P^{(t)}_1, t) 
- \frac{1}{2} \epsilon_{\pi_2}(P^{(t)}_1, t)+ \epsilon_{\mu_2}(L^{(t)}_1, R^{(t)}_1, t)
\end{gather*}
Such a representation leads to a cumulative score calculation of the joint distribution of all the nodes of interest to the factor using linear addition and subtraction. 
We can realize from~\autoref{eq:gfc_prob_model} that the final score function depends on the composition of all the factors in the spatial-temporal factor graph. While factors $f\in\mathcal{F}$ are mostly modeled implicitly by the temporal skills, the external factors can be any arbitrary spatial constraints that ensure the satisfaction of the pre-condition of the subsequent skills. Hence, with new additions to the set of external factors $\mu'\in\mathcal{M'}$, one can reuse the same temporal skills with added new spatial constraints.

\textbf{Summary} \Ours is a new paradigm to solve complex manipulation problems using spatial-temporal factor graphs. GFC can be divided into the following segments: (1) train individual skill factor distributions individually, without any prior knowledge or data from other skills in the library 
(2) create spatial-temporal factor graph from a plan skeleton, (3) compose individual spatial and temporal factor distributions to construct a probabilistic graphical model, and (4) use the plan-level distribution to sample plan solutions. The proposed approach is modular as the individual skill factors and constraints can be flexibly connected to form new graphs. GFC can connect parallel skill chains with added spatial factors to solve coordinated manipulation problems directly at inference. Additional detail in~\autoref{algo:gfc}.

\section{Experiment}

In this section, we seek to validate the following hypotheses: (1) GFC relaxes strict temporal dependency to allow spatial-temporal reasoning, performing better or on par with prior works in single-arm long-horizon sequential manipulation tasks, (2) GFC can effectively solve unseen coordination tasks, and (3) GFC is adept in reasoning about long-horizon action dependency while being robust to increasing task horizons. 
We systematically evaluated our method on 9 long-horizon single-arm manipulation tasks from prior works and 4 complex multi-arm coordination tasks in simulation. We also demonstrate deploying \Ours on a bimanual Franka Panda setup in the real world.

\textbf{Relevant baselines and metrics:} Our proposed method is based on factorized states and supports long-horizon planning for collaborative tasks directly at inference via probabilistic chaining. In this context, we consider prior methods based on probabilistic chaining with vectorized states~(\textbf{GSC}~\cite{mishra2023generative}) and discriminative search-based approaches for solving long-horizon planning by skill chaining: with uniform priors~(\textbf{Random CEM} or \textbf{RCEM}) or learned policy priors~(\textbf{STAP}~\cite{agia2022taps}). Since all prior works use sequential planning, we compare the performance of the proposed method on the sequential version of the parallel skeleton.

\subsection{Setup}

\textbf{Skill Data Collection and Skill Training} We consider a finite set of parameterized skills in our skill library. While our framework supports flexible addition of new skills to the skill library, we choose skills appropriate for the considered tasks. The parameterization, data collection, and training method for each of the skills is described as follows:

\begin{enumerate}
    \item \texttt{Pick}:~Gripper picks up an object from the table and the parameters contain 6-DoF pose in the object's frame of reference. The skill diffusion models are trained on successful pick actions on all the available set of objects namely lid, cube, hammer, and nail/stake. 
    \item \texttt{Place}:~Gripper places an object at the target location and parameters contain 6-DoF pose in the place target's frame of reference. This skill requires specifying two set of parameters, the target pose and the target object (e.g. box, table). The picked object is placed and successful placements are used to train the skill diffusion model.
    \item \texttt{Move}:~Gripper reaches a target location with an object in hand and parameters contain 6-DoF pose in the manipulator's frame of reference within the workspace. This skill captures the distribution of the reachable workspace of the robot. When composed with the \texttt{Move} skill of the second manipulator, the combined distribution captures the common workspace.
    \item \texttt{ReGrasp}:~Gripper grasps object mid-air and the parameters contain 6-DoF pose in the object's frame of reference. While collecting data directly for this skill is non-trivial, we consider that if an object is picked up with parameters $q_1$ and moved with parameters $q_2$, then the object can be grasped at the workspace location defined by $q_2$ with the \texttt{ReGrasp} parameters as $q_1$. Thus, we reuse \texttt{Pick} and \texttt{Move} data to train the skill diffusion model for \texttt{ReGrasp}. While this is a design choice, with appropriate skill level data, we can train this skill separately too.
    \item \texttt{Push}:~Gripper uses the grasped object to push away another object. The skill is motivated from prior work~\cite{mishra2023generative, agia2022taps} where a hook object is used to \texttt{Push} blocks. The parameters of this skill are $(x, y, r, \theta)$ such that the hook is placed at the $(x, y)$ position on the table and pushed by a distance $r$ in the radial direction $\theta$ w.r.t. the origin of the manipulator. The skill diffusion models is trained following GSC~\cite{mishra2023generative}.
    \item \texttt{Pull}:~Gripper uses the grasped object to pull another object inwards. The skill is also motivated from prior work~\cite{mishra2023generative, agia2022taps} where a hook object is used to \texttt{Pull} blocks. The parameters of this skill are $(x, y, r, \theta)$ such that the hook is placed at the $(x, y)$ position on the table and pulled by a distance $r$ in the radial direction $\theta$ w.r.t. the origin of the manipulator. The skill diffusion models is trained following GSC~\cite{mishra2023generative}.
    \item \texttt{Strike}:~Gripper strikes another object with one object in hand (e.g., a hammer). As a design choice, we do not train a skill diffusion model for this skill. \texttt{Strike} is primarily used as a terminal skill. We are only concerned about the pre-condition as their effects can be designed manually, which is similar to “subgoal skill” used in prior work. For example, in order to satisfy the pre-condition of \texttt{Strike}, the hammer and nail must be aligned. This can be satisfied in diverse configurations. However, the effect is achieved through a deterministic motion.
    \item \texttt{Pour}:~Gripper rotates the object in hand in a pouring fashion. Similar to \texttt{Strike}, we use \texttt{Pour} as a terminal skill too. In order to satisfy the pre-condition of \texttt{Pour}, the transform between the source and target mug must belong to the family of admissible distributions. We achieve the actual trajectory by designing a deterministic motion. With appropriate skill level data, we can also train skill diffusion models, however, such improvement is out of scope of this work.
\end{enumerate}

\textbf{Training.} We train individual skill diffusion score-functions using the denoising score-matching~(DSM) loss following~\autoref{algo:training}. We collect datasets of transitions observed during the execution of a skill on an object and use them to train the score networks. The dataset size varies according to the difficulty and diversity of a skill's execution on a particular object. For example, we need 100 successful \texttt{Pick} parameters for training the skill to pick the hammer and 300 successful \texttt{Move} parameters to cover the whole workspace of the robot. For \texttt{ReGrasp}, we use both the \texttt{Pick} and \texttt{Move} parameters.

\textbf{Effect of training data coverage.} If we consider ``ideal" score functions and a perfect representation of the factor distributions, a solution exists if there is an overlap between two connected factor distributions. If such an overlapping segment does not exist, GFC will not be able to complete the spatial-temporal plan. Hence, the training data for each factor (here temporal factors only) must be diverse enough to ensure that the overlap exists. For example, a successful handover in \textit{Hammer Place} and \textit{Hammer Strike} is not possible if the training data only consists of \texttt{Pick} parameters to pick the hammer from the center of the handle. Similarly, if the training data for \texttt{Move} does not cover the common workspace of both robots, our proposed algorithm will be unable to complete the coordinated plan.

\textbf{Example of spatial factors.} Previous work~\cite{yang2023compositional} considered a family of spatial factors like (\texttt{left}, \texttt{right}, \texttt{top}, \texttt{bottom}, \texttt{near} and \texttt{far}) to model collision-free object configurations. In this work, we are particularly interested in constructing a family of fixed transforms (\texttt{FixedTransform}) to model coordinated manipulation motion. For example, in order to satisfy the pre-condition of \texttt{strike(A, B)}, the transform between nodes $A$ and $B$ must satisfy a family of transforms signifying that $B$ must be \texttt{Aligned} with  $A$ to strike it. Thus the factor for \texttt{strike(A, B)} with \texttt{Aligned} transforms~$\mathcal{H}_A$ will look like: $f \equiv \text{distance}(\text{transform}(A,B), \mathcal{H}_A) \leq \text{permisible error}$ for at least one transform. In that case, the distribution of the factor will be: {$p(f=\texttt{True} |A,B) \propto \exp[-\text{distance}(\text{transform}(A,B), \mathcal{H}_A)]$}. The score of such a distribution can then be calculated as 
\[
\epsilon_f(A^{(t)}, B^{(t)}, t) = -\nabla_{A^{(t)}, B^{(t)}} \text{distance}(\text{transform}(A^{(t)}, B^{(t)}), h_A)
\]
where $h_A \in \mathcal{H}_A$ is the closest transform to the current transform. The distance between transforms is calculated as the summation of the Cartesian distance and the quaternion distance.

\begin{figure}[t]
    \centering
    \includegraphics[width=\linewidth]{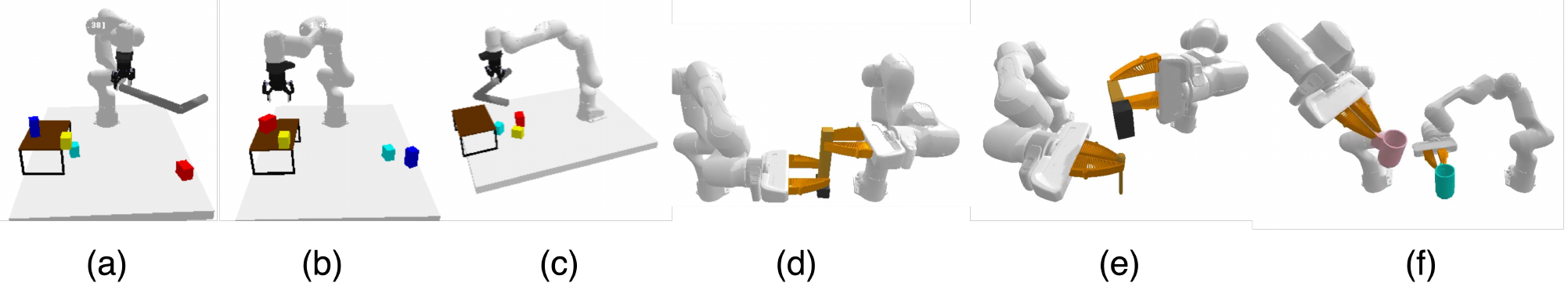}
    \caption{Evaluation tasks: \textbf{(a) Hook reach:} Hook is used to pull an object in the robot's workspace followed by other skills. \textbf{(b) Constrained packing:} Multiple objects must be placed on a rack without collisions. \textbf{(c) Rearrangement push:} Hook is used to push objects to a desired arrangement followed by other skills. \textbf{(d) Hammer place}: A hammer must be handed over to another manipulator and placed in a target box. \textbf{(e) Hammer nail}: A hammer must be handed over to another manipulator and a configuration must be achieved to strike a nail. \textbf{(f) Pour cup:} Cups must be brought in a configuration that allows successful pouring from one to another. }
    \label{fig:all_tasks_chain}
\end{figure}

\subsection{Key Findings}

\textbf{GFC relaxes strict linear dependency assumptions.} We first evaluate GFC on single-manipulator long-horizon tasks introduced by STAP~\cite{agia2022taps} and also used by GSC~\cite{mishra2023generative}. These tasks consider manipulation by reasoning about the usage of a tool~(a hook) to manipulate blocks out of or into the robot workspace (sample initial states shown in \autoref{fig:all_tasks_chain}(a-c)). \textit{Hook Reach} is to hook the cube in order for the arm to grasp and move the block to a target. \textit{Rearrangement Push} requires placing a cube such that it can be pushed beneath a rack using the tool. \textit{Constrained Packing} is to place four cubes on a constrained surface without collisions. While these tasks are originally designed to highlight linear sequential dependencies, there are steps with indirect dependencies or independence that only \Ours can effectively model because of the factorized states. For example, in \textit{Rearrangement Push}, the picking pose of the cube should not affect the tool use steps. As shown in~\autoref{tab:results_sequential}, we observe that the performance of GFC is consistently on-par with the baseline for tasks with strict linear dependencies such as \textit{Hook Reach} and on-par or better for tasks with more complex dependency structures such as \textit{Rearrangement Push}. This validates our hypothesis that \Ours effectively models sequential dependencies, in addition to independence and skipped-step dependencies in long-horizon tasks.

\begin{table*}[t]
    \centering
    \caption{We show performance comparison of our method with relevant baselines on 9 single manipulator tasks and 3 two-manipulator tasks based on 100 trials for each of them. The task length shows the relative difficulty of solving them. We also conduct evaluation on 3 extended tasks to show robustness of \Ours to task length~($|\mathcal{T}|$) and efficient reasoning about interstep dependencies. More details about the environments are provided in~\autoref{app:evaluation-tasks} and~\autoref{app:hammer-nail-extensions}. We also provide the breakdown of success rate per skill step in~\autoref{sec:task-breakdowns}.}
    \vspace{0.2cm}
    \resizebox{\textwidth}{!}{
    \begin{tabularx}{\textwidth} { 
    | c
    | c 
    | c
    | c
      | c
      | c
      | c
      | c 
      | c |}
     \cline{1-9}
       \multicolumn{3}{|c|}{Evaluation Tasks} & RCEM & DAF~\cite{xu2021deep} & STAP~\cite{agia2022taps}  & GSC~\cite{mishra2023generative}  & GFC  & $|\mathcal{T}|$ \\
       \cline{1-9}
       \multirow{9}*{\rotatebox[origin=c]{0}{\makecell{Single \\ Manipulator}}} & \multirow{3}*{Hook Reach} & T1 & 0.54 & 0.32 & \textbf{0.88} & 0.84 & 0.82 & 4 \\
         & & T2 & 0.40 & 0.05 & 0.82 & \textbf{0.84} & 0.82 & 5 \\
         & & T3 & 0.30 & 0.00 & 0.76 & 0.76 & \textbf{0.80} & 5 \\
       \cline{2-9}
        & \multirow{3}*{\makecell{Rearrangement \\ Push}} & T1 & 0.30 & 0.0 & 0.40 & \textbf{0.68} & \textbf{0.68} & 4\\
         & & T2 & 0.10 & 0.08 & 0.52 & 0.60 & \textbf{0.65} & 6\\
         & & T3 & 0.02 & 0.0 & 0.18 & 0.18 & \textbf{0.25} & 8\\
       \cline{2-9}
        & \multirow{3}*{\makecell{Constrained \\ Packing}} & T1 & 0.45 & 0.45 & 0.65 & \textbf{0.75} & \textbf{0.75} & 6\\
         & & T2 & 0.45 & 0.70 & 0.68 & \textbf{1.0} & \textbf{1.0} & 6 \\
         & & T3 & 0.10 & 0.0 & 0.20 & \textbf{1.0} & \textbf{1.0} & 8\\
       \cline{1-9}
       \multirow{3}*{\rotatebox[origin=c]{0}{\makecell{Bimanual \\ Manipulation}}} & \multicolumn{2}{|c|}{Hammer Place} & 0.05 & - & 0.28 & 0.41 & \textbf{0.63} & 8 \\
        & \multicolumn{2}{|c|}{Pour Cup} & 0.10 & - & 0.18 & 0.15 & \textbf{0.41} & 4 \\
        & \multicolumn{2}{|c|}{Hammer Nail} & 0.02 & - & 0.15 & 0.15 & \textbf{0.34} & 11 \\
        \cline{1-9}
        \multicolumn{9}{|c|}{Longer Horizon Evaluation Tasks} \\
        \cline{1-9}
        \multicolumn{7}{|c|}{Handback Hammer Nail} & \textbf{0.24} & 16 \\
        \multicolumn{7}{|c|}{Handback Hammer Nail w/ auxilliary tasks} & \textbf{0.25} & 18 \\
        \multicolumn{7}{|c|}{Handback Hammer Nail w/ extended auxilliary tasks} & \textbf{0.21} & 20 \\
       \cline{1-9}
    \end{tabularx}
    }
    \label{tab:results_sequential}
\end{table*}

\textbf{GFC can solve complex coordinated manipulation tasks.} Here, we aim to validate that \Ours can effectively plan and solve different types of coordinated manipulation tasks. We present results on tasks with increased collaboration challenges. First, we consider tasks that require coordination but can be serialized into interleaved skill chains and solved by prior skill-chaining methods. \textit{Hammer Place}, as shown in~\autoref{fig:hammer-variants}, is for one arm to pick a hammer, hand it over to another arm for placemement into a target box. \textit{Hammer Nail} is an extension where, after hammer handover, first arm picks up a nail and both arms coordinate to move to positions such that the hammer's head is aligned with the nail for the subsequent striking step. The task is illustrated in~\autoref{fig:hammer-variants}. As evident from \autoref{tab:results_sequential}, \Ours significantly outperforms all baselines in both tasks. The gap is larger in the more challenging \emph{Hammer Nail} task, which includes additional spatial and temporal constraints such as the hammer must be re-grasped towards the tail end for the subsequent hammering step, and the hammer and nail must be aligned for a successful strike. This demonstrates that \Ours can effectively model and resolve both spatial and temporal constraints in complex tasks. 

\begin{wrapfigure}{r}{0.35\linewidth}
\includegraphics[width=\linewidth, clip, trim={0.2cm 0.26cm 0.2cm 0.5cm}]{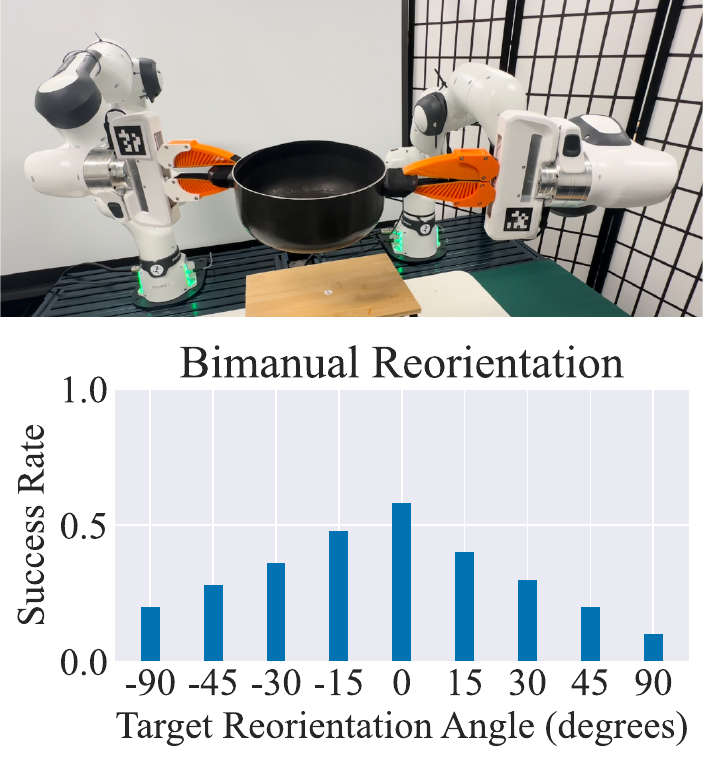}
    \caption{Evaluating \Ours on bimanual reorientation where two arms simultaneously pick and reorient a pot.}
    \label{fig:bimanual_success_real}
\end{wrapfigure}
\textbf{\Ours can zero-shot generalize to new bimanual tasks by composing single-arm skill chains.} 
The \textit{Pour Cup}~(\autoref{fig:pour_cup_chain_hardware}) task is to \texttt{Pick} a cup with each arm, \texttt{Move} to position the two cups, and \texttt{Pour} the content of one into the other. \Ours can directly reuse \texttt{Pick} and \texttt{Move} skill models and adapt the \texttt{Strike} skill model for the \texttt{Pour} step by adding a new spatial constraint. Unlike hammer that can strike from either face of the head, the cups can only be poured using the open top and not the closed bottom. The constraint can be directly added as a factor and optimized globally through guided diffusion process. A quantitative comparison is shown in~\autoref{tab:results_sequential}.

Finally, we consider the \textit{Bimanual Reorientation}~(\autoref{fig:bimanual_reorientation_chain_hardware}) task where two arms must simultaneously operate on the same object of interest~(a pot), lift it up, and rotate it to a target reorientation angle~(about z-axis) as illustrated in~\autoref{fig:bimanual_success_real}~(Top) for a $45$-deg angle. The tasks must be solved via parallel skill chaining with spatial constraints and hence none of the prior baselines can be used. The factor graph~(~\autoref{fig:parallel_chaining_example} Right) includes a spatial fixed transform constraint between both the arms and hence the subsequent skills operate while satisfying the constraint. 
\autoref{fig:bimanual_success_real}~(Bottom) shows a detailed task success rate breakdown given different orientation goals. The spatial and temporal challenges posed by the task are detailed further in~\autoref{app:evaluation-tasks}.

\begin{figure}[t]
    \centering
    \includegraphics[width=\linewidth]{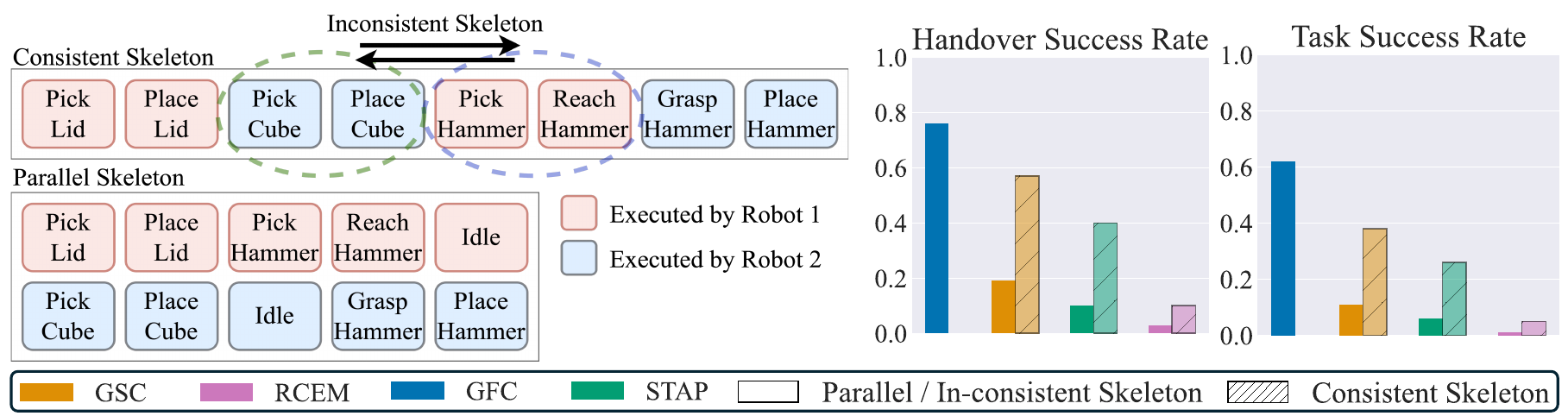}
    \caption{\textbf{Linear chaining has limitations.} Baseline methods with linear chain assumption suffers from performance drop when given inconsistent skill chains, where steps with sequential dependencies are swapped. \Ours retains high success rate using the parallel skeleton.}
    \label{fig:inconsistent_consistent}
\end{figure}

\textbf{\Ours can handle independence and inconsistent skill chains.} Here, we analyze how \emph{independent} steps in a sequential manipulation chain affects the performance of each method. We consider \textit{Hammer Place}, where the order of transporting the cube and handing over hammer is interchangeable. As illustrated in~\autoref{fig:inconsistent_consistent}, we consider a \emph{consistent} plan skeleton where sequentially-dependent steps for the two main objectives, i.e., (1) opening lid then transporting cube and (2) picking, handing over, and placing hammers, are completely sequentially. We also consider an \emph{inconsistent} plan skeleton where the steps are interleaved.  
We show the handover success and overall task success in~\autoref{fig:inconsistent_consistent}~(Right). A successful handover requires choosing compatible parameters for \texttt{Pick}, \texttt{Regrasp}, and \texttt{Move} skills. While this increases the difficulty leading to lower scores in the handover success rate, even with a minor distraction in \textit{inconsistent} skeleton, the previous approaches failed to propagate the skipped-step dependencies as evident from the task success rate.



\begin{figure}[t]
\centering
    \includegraphics[width=\linewidth]{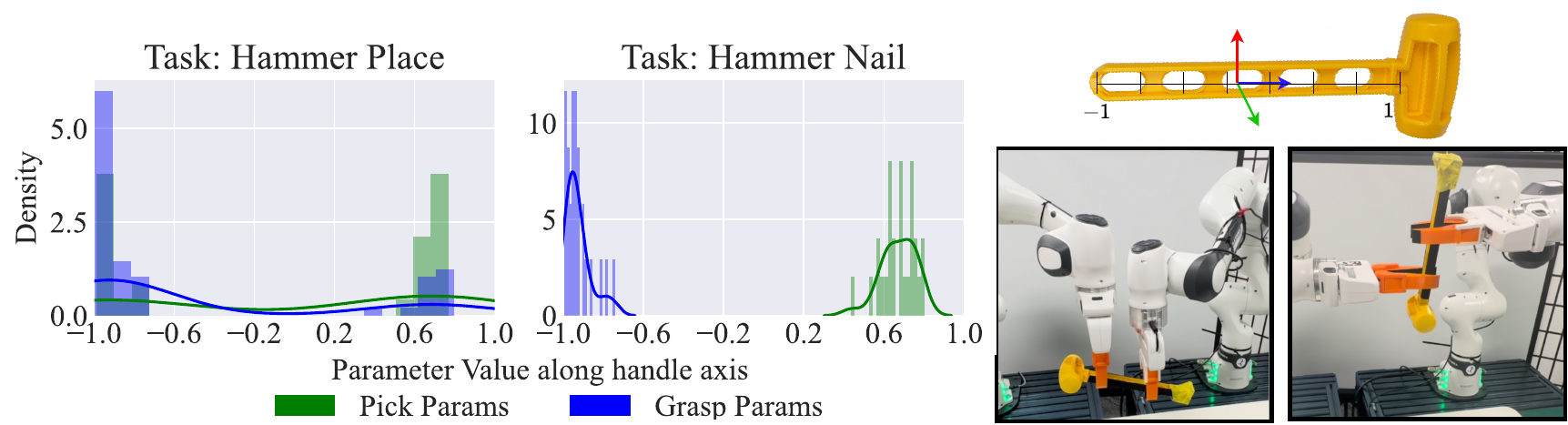}
    \caption{\textbf{Analysis of coordination.} We show that the planner is able to reason about the long-horizon action dependency of \texttt{Pick} and \texttt{Grasp} skills. (Left) While we see that \textit{Hammer Place} can be solved by pick/grasp at head/tail and vice versa, to satisfy the precondition of \texttt{Strike} in \textit{Hammer Nail}, the hammer must be grasped near tail so must be picked near head. (Right) We show orientation reasoning, where the hammer can either be grasped on the same side or the flip side.}
    \label{fig:hammer_grasping}
\end{figure}

\begin{figure}[h]
    \centering
    \includegraphics[width=1\linewidth]{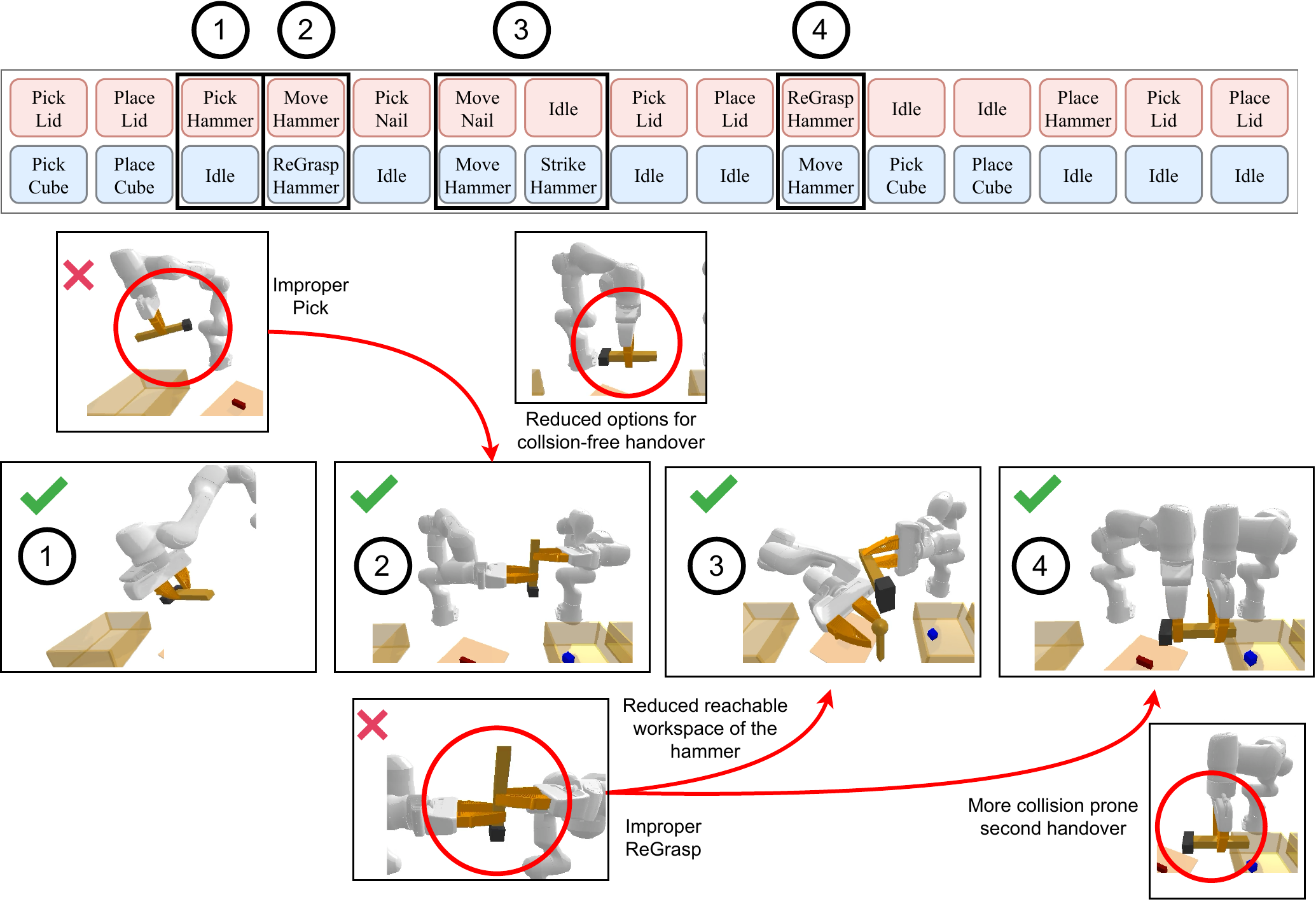}
    \caption{\textbf{Inter-step dependencies.} We show the steps and reasoning required to solve the \textit{Hammer Nail} task. An improper initial pick can lead to a failed or unfavorable handover which might lead to difficulty in performing \texttt{Strike} and the second handover. Thus the algorithm must reason about inter-step action dependency over longer horizons to solve the task successfully.}
    \label{fig:inter-step}
\end{figure}

\textbf{GFC can reason about action dependency while being robust to increasing task horizons.} We observe in ~\autoref{fig:hammer_grasping}~(left) that while \textit{Hammer Place} task can be solved by picking or grasping on any end of the hammer handle, \textit{Hammer Nail} requires more constrained parameter sampling. Further, in addition to the parameter selection along the handle axis, the method also samples suitable orientation~(same or flip side) for grasping as shown by two examples in~\autoref{fig:hammer_grasping}~(right). We further give an example of the capability of our method in handling longer horizon inter-step dependencies.
We particularly want to emphasize that hammering a nail not only requires extensive affordance planning to perform a handover but also requires allowing sufficient reachable workspace to align the hammer head with the nail. For an even longer task which requires performing a second handover, the choice of parameters for the first handover also affects the success of the second handover, thus increasing the action-dependency horizon as illustrated in~\autoref{fig:inter-step}. Our framework is able to compose learned factors~(diffusion models) to solve a wide variety of tasks, as long as their solutions fall in the combinatorial space. We also want to emphasize that GFC is robust with respect to the task length as shown in~\autoref{tab:results_sequential}.

\begin{wrapfigure}{r}{0.36\linewidth}
\includegraphics[width=\linewidth]{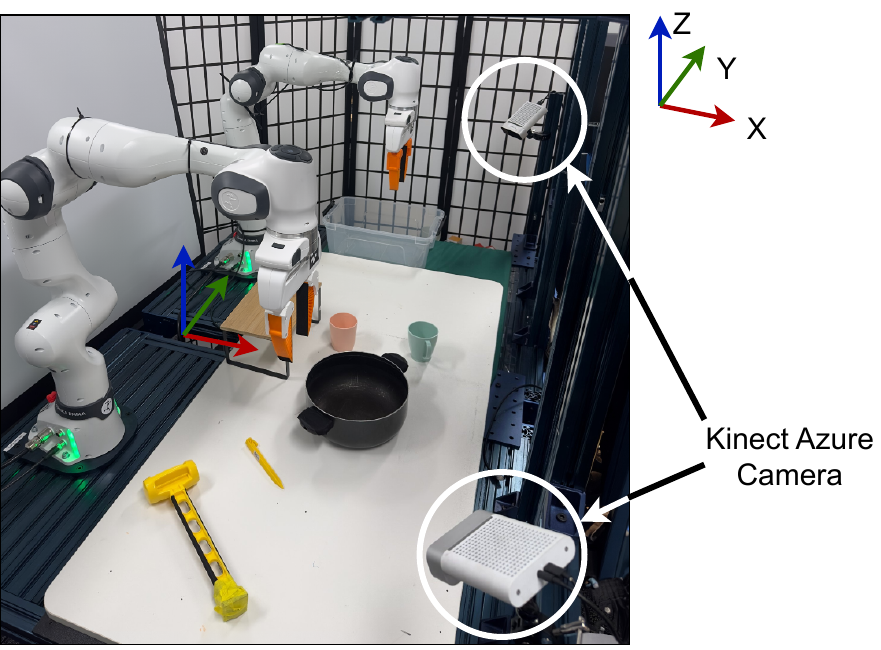}
\caption{Real-World Experimental Setup
}
\label{fig:real_robot_setup_v2}
\end{wrapfigure} 

\subsection{Real Robot Experiments}
\label{app:real-setup}

\textbf{Complete setup.}
We use two Franka Panda robot arms placed in parallel to demonstrate the coordinated tasks as illustrated in~\autoref{fig:real_robot_setup_v2}. A pair of flexible Finray fingers~\cite{crooks2016fin} is attached to the parallel jaw grippers. For each of the arm, we set up a Kinect Azure camera calibrated to the origin of the arm. We use objects like mallet~(hammer), stake~(tent peg, nail), garden foam, a kitchen pot, two types of mugs and a rack for the considered tasks. We use segment-anything~\cite{kirillov2023segany} and CLIP~\cite{radford2021learning} to segment the objects from the RGBD image based on text descriptions and use the segmented masks to obtain the point clouds for the objects. Finally, we use ICP to align the obtained and model point clouds to calculate the transformation of the object. The procedure is done for both cameras to obtain transforms for all the detected objects in both robot's frame of reference. For a particular object, we select the transform from the arm closest to the object to get precise pose estimation~(due to better depth data). We finally use the obtained transforms to recreate the physical scene in simulation, employ GFC in simulation and rollout the results in the real-world. While planning, the Frankx controller~\cite{frankx} is used to generate smooth motion toward the desired pose.

\begin{figure}[h!]
    \centering
    \includegraphics[width=0.8\linewidth, trim={0, 0.2cm, 0, 0}, clip]{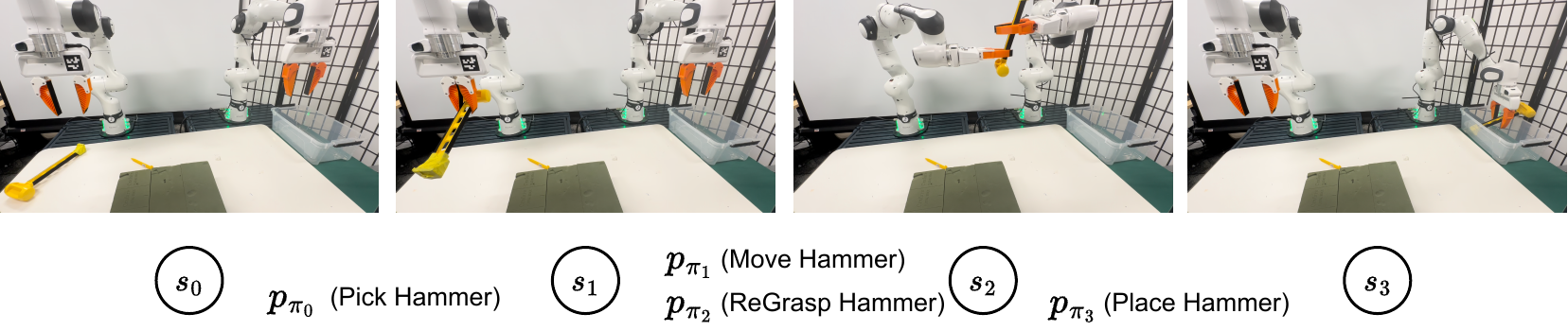}
    \caption{\textbf{Coordination task: \textit{Hammer Place}} The left arm must handover the hammer to the right arm such that the hammer can be placed inside the box.}
    \label{fig:hammer_place_chain_hardware}
\end{figure}

\begin{figure}[h!]
    \centering
    \includegraphics[width=\linewidth, trim={0, 0.2cm, 0, 0}, clip]{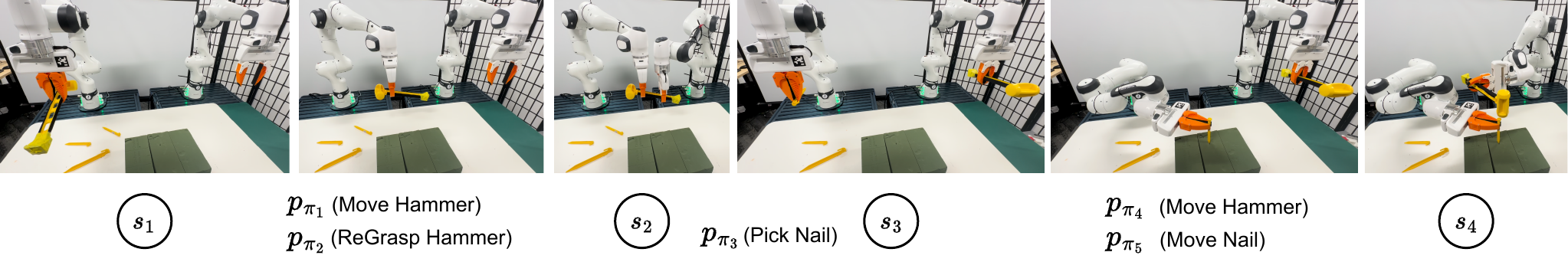}
    \caption{\textbf{Coordination task: \textit{Hammer Nail}} The left arm must handover the hammer to the right arm and pick up the nail. Both arms have to coordinate in order to move the hammer and nail to a configuration in which the hammer can strike the nail.}
    \label{fig:hammer_nail_chain_hardware}
\end{figure}

\begin{figure}[h!]
    \centering
    \includegraphics[width=\linewidth, trim={0, 0.2cm, 0, 0}, clip]{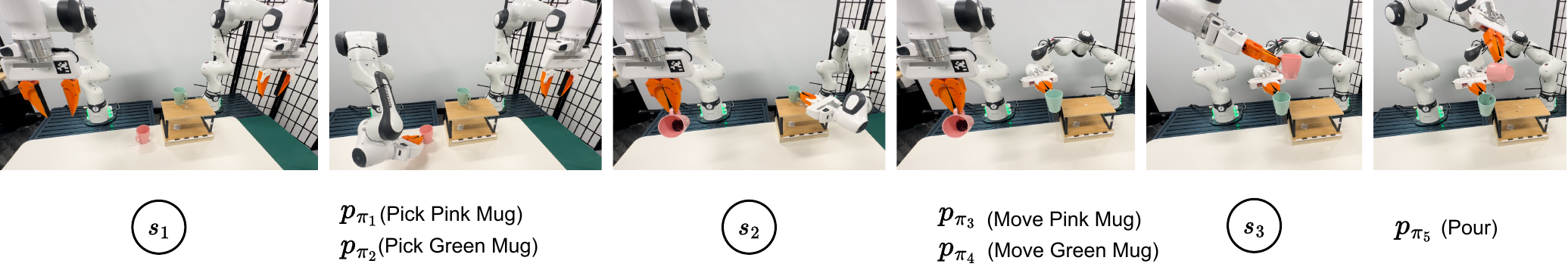}
    \caption{\textbf{Coordination task: \textit{Pour Cup}} The left arm and right arm must pick up the pink mug and green mug respectively. Both arms have to coordinate in order to move the mugs to a configuration in which the left arm can pour the pink mug contents into the green mug.}
    \label{fig:pour_cup_chain_hardware}
\end{figure}

\textbf{Qualitative analysis.} We perform qualitative analysis for all four coordinated tasks using the hardware setup as shown in~\autoref{fig:hammer_place_chain_hardware},\autoref{fig:hammer_nail_chain_hardware},~\autoref{fig:pour_cup_chain_hardware} and~\autoref{fig:bimanual_reorientation_chain_hardware}. We further provide detailed videos of execution in the supplementary video.

\begin{figure}[h!]
    \centering
    \includegraphics[width=0.6\linewidth]{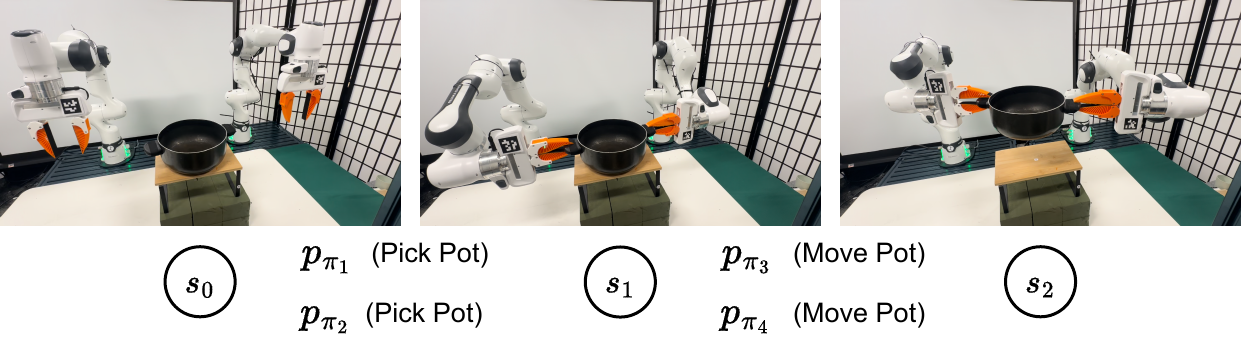}
    \caption{\textbf{Coordination task: \textit{Bimanual Reorientation}} The left arm and right arm must pick up the pot simultaneously. Both arms have to coordinate in order to rotate the pot to a specified target reorientation angle. For the above illustration, the reorientation angle is 30deg.}
    \label{fig:bimanual_reorientation_chain_hardware}
\end{figure}

\textbf{Failure analysis.} We try to analyze the reason for the failure of GFC in certain cases. A limiting factor of our planning framework is that the nodes denote waypoints required to be reached for completing the geometric execution and satisfying the goal condition without caring about the trajectory between them. Since we do not explicitly provide the intuition of inverse kinematics~(IK) or collision, we assume that these properties are learned implicitly using the successful transitions in the training data. Hence, apart from sim-to-real gap~(consisting of pose-estimation error, nature of surfaces in contact, and weight of the objects like hammer and pot), the primary reasons for failure are: (1) sampling a pose where IK cannot be computed, i.e. unreachable. (2) The sampled pose is not collision-free. We provide sim-to-real gap failures in the supplementary video.

\section{Limitations and Future Directions} 
\label{sec:limitation}

First, our method does not generate high-level task plans. Solving the full TAMP problem with a unified generative model is an important future direction. 
Second, our method operates in a low-dimensional state space \um{and hence requires a state estimator}. We plan to extend \Ours to work with high-dimensional observations. Finally, similar to prior works~\cite{xu2021deep,agia2022taps,mishra2023generative}, our approach operates on parameterized skills. Future work can explore integrating learned skills or trajectory generators for additional generality and scalability.

\section{Conclusion} 
\label{sec:conclusion}
We presented \Ours, a learning-to-plan method for complex coordinated manipulation tasks. \Ours can flexibly represent  multi-arm manipulation with one or more objects with a spatial-temporal factor graph. During inference, \Ours composes factor graphs where each factor is a diffusion model and samples long-horizon plans with reverse denoising. \Ours is shown to solve sequential and coordinated tasks directly at inference and reason about long-horizon action dependency across multiple temporal chains. Our framework generalizes well to unseen multiple-manipulator tasks.


\bibliography{references}  

\newpage
\appendix

\renewcommand{\theequation}{S\arabic{equation}}
\renewcommand{\thesection}{S\arabic{section}}
\renewcommand{\thefigure}{S\arabic{figure}}
\renewcommand{\thetable}{S\arabic{table}}

\section{Main Contributions}
\label{app:main-contributions}

\textbf{Generative Factor Chaining~(GFC)} is proposed with the motivation of zero-shot motion planning for long-horizon tasks. The goal is to use short-horizon skill transition distributions and efficiently compose them to structure a long-horizon task-level distribution at inference. The factorized state representation of GFC allows explicit reasoning of inter-object and skill-object interactions and satisfying spatial constraints for coordinate manipulation.
The primary contributions of GFC are as follows:
\begin{enumerate}
    \item A \textbf{generalized task representation} to formulate complex long-horizon coordination tasks as a spatial-temporal factor graph of single-arm manipulation skill sequences connected via spatial dependencies.
    \item A \textbf{compositional framework} to compose short-horizon skill-level transition distributions learned via diffusion models to represent long-horizon task-level distributions.
    \item \textbf{Easy plug-and-play} via learning skill distributions with skill-level data only and add it to the skill library. Any skill from the library can be plugged as temporal factors in the spatial-temporal factor graph directly at inference for a given long-horizon task.
\end{enumerate}

\section{Additional Related Works}

\textbf{Factor-graph representation for TAMP.} The graphical abstraction of a system for understanding several inter-dependencies has been used in various domains~\cite{Dellaert2021FactorGE}. Specifically in context to task and motion planning~(TAMP), such a representation allows the decomposition of multiple modalities~(discrete and continuous variables) in the state of a system~\cite{garrett2021integrated}. Solving together for discrete~(logical decision variables) and continuous~(motion parameters) can be formulated as a Hybrid Constraint Satisfaction Problem (H-CSP) problem, Logic-Geometric Program (LGP)~\cite{toussaint2015logic}, and more recently by advanced gradient descent methods~\cite{lee2023stamp}. By following the factor-graph representation, the state space can be represented as a Cartesian product of all the subspaces and the action space can be compactly represented based on the modalities they affect. We particularly follow the \emph{dynamic factor graph} representation used by \citet{garrett2021integrated} to represent all the objects and action parameters as the variable nodes of the graph and all the kinematic inter-dependencies as the factors of the graph.

\textbf{Optimization for factor graphs.} Factor graphs are graphical models where the directed and undirected factors, respectively represent the joint or conditional distribution of the variable nodes connected to them. Most directed factors graphs as used for localization~\cite{gtsam, xiao2024multi} are formulated into probabilistic graphical models of hidden-markov chains and solved for the maximum a posteriori~(MAP)~\cite{gtsam, hao2023blitzcrank} estimates of the unknown node variables. Particularly in motion planning, optimizing for all the variable nodes is often formulated as a constraint satisfaction problem~\cite{garrett2021integrated, yang2023compositional}. 

\textbf{Additional related works on learning for TAMP.} Recent works have shown that a number of components of a TAMP system benefit from powerful generative models. Wang et al~\cite{wang2018active,wang2021learning} use Gaussian Processes to learn continuous-space sampler for TAMP. Similarly, Kim et al.~\cite{kim2017guiding} use GANs to learn action samplers. Fang et al.~\cite{fang2023dimsam} propose to use Diffusion Models to capture complex distributions such as Inverse Kinematics solutions, grasps, and contact dynamics. However, they still rely on an overarching TAMP system to consume the generated samples to perform planning. In contrast, our method directly forms a geometric plan sampler by chaining together factor-level diffusion models. 

\newpage
\begin{algorithm*}[h]
    \caption{Generative Factor Chaining~(GFC) Algorithm}
    \SetAlgoLined
    \label{algo:gfc}
    $\text{ \textbf{Hyperparameters:} }$ \\
    $\text{ Number of reverse diffusion steps } T$ \\
    \vspace{1em}
    $\text{ \textbf{Inputs}: }$\\
    $\text{ Pre-defined skill library } \Pi = \{\pi_1, \pi_2, \dots, \pi_M\}$ \\
    $\text{ Individual skill diffusion score functions } \epsilon_\pi$ \\
    $\text{ Task skeleton } \Phi_K~=~\{\pi_0, \pi_1, \dots, \pi_K\} \text{: a sequence of skills of length } K$ \\
    $\text{ Scene graph sequence } \Phi_S~=~\{s_0, s_1, \dots, s_K\} \text{: a sequence of scene factors of length } K+1 \text{ where } s_k~\equiv~\{\mathcal{V}_k, \mathcal{F}_k\}$ \\
    $\text{ Goal condition } g \equiv \{\mathcal{V}_g, \mathcal{F}_g\}$ \\
    $\text{ Noise schedule } \sigma$ \\
    \vspace{1em}

    $\text{ Initialize } t = T = 1$ \\
    $\text{ Initialize } \Delta t$ \\
    $\text{ Initial node sequence } \textbf{x}^{(T)} = \Big[v^{(T)}_k \;\forall\; v \in \mathcal{V}_k, \;\;a^{(T)}_{\pi_k}, \dots \forall \; k \in [0, K]\Big] \text{ sampled from } \mathcal{N}(\textbf{0}, \sigma_T\textbf{I})$\\
    \vspace{1em}
    \While{$t \geq 0$}{
        \vspace{1em}
        $\text{ // Score of the joint distribution of all the nodes}$ \\
        $\epsilon_{\Phi}(v^{(t)}_k \;\forall\; v \in \mathcal{V}_k, \;\;a^{(t)}_{\pi_k}, \dots \forall \; k \in [0, K], t) = \textbf{0}$ \\
        \vspace{1em}
        $\text{ // Calculating the effective score of each node}$ \\
        $\epsilon_{\Phi}(x^{(t)}, t) = \sum^K_{k=0} \epsilon_{\pi_k}(x^{(t)}, t) + \sum^K_{k=0} \sum_{f \in \mathcal{F}_k} \epsilon_{f}(x^{(t)}, t) \;\;\; \forall x \in \textbf{x} \;\; \text{ (Computational assumption, \autoref{eq:final_gfc})}$

        \vspace{1em}
        $\text{ // Only for nodes connected with two temporal factors } f_{x,1} \text{ and } f_{x,2}$ \\
        $\epsilon_{\Phi}(x^{(t)}, t) = \epsilon_{\Phi}(x^{(t)}, t) - \frac{1}{2}\Big[ \epsilon_{f_{x,1}}(x^{(t)}, t) + \epsilon_{f_{x,2}}(x^{(t)}, t) \Big] \quad \text{ (Denominator compensation, \autoref{eq:final_gfc})}$
        
        \vspace{1em}
        $\text{ // calculating updated noised samples for the next reverse diffusion timestep}$ \\
        $\tilde{\textbf{x}}^{(t-1)} = \textbf{x}^{(t)} + \dot{\sigma}_t\sigma_t \epsilon_{\Phi}(v^{(t)}_k \;\forall\; v \in \mathcal{V}_k, \;\;a^{(t)}_{\pi_k}, \dots \forall \; k \in [0, K], t)\Delta t$ \\
        $t= t - \Delta t$\\
    }
    $\text{Return } \textbf{x}^{(0)} $
\end{algorithm*}


\newpage

\vspace{10pt}
\section{Model Training and Architecture}
\label{app:model-training}

\textbf{Model architecture.} Our transformer-based score-network architecture is derived from the Diffusion Models with Transformers~(DiT)~\cite{Peebles2022DiT} implementation, also open-sourced at: \url{https://github.com/facebookresearch/DiT}. We follow a similar concept to that of patchifying an image into many smaller patches, encoding each one of them using a common encoder and passing it as a sequence to the transformer architecture with respective positional embeddings. In our case, we consider a sequence of nodes consisting of both the object and skill parameters nodes in the factor graph as the input sequence. Each node variable is encoded into a common dimension using a common object node encoder and skill parameter encoder for object and skill parameter nodes respectively. The output is decoded into their respective dimensions using similar decoder setup.

\begin{figure}[h]
    \centering
    \includegraphics[width=0.9\linewidth]{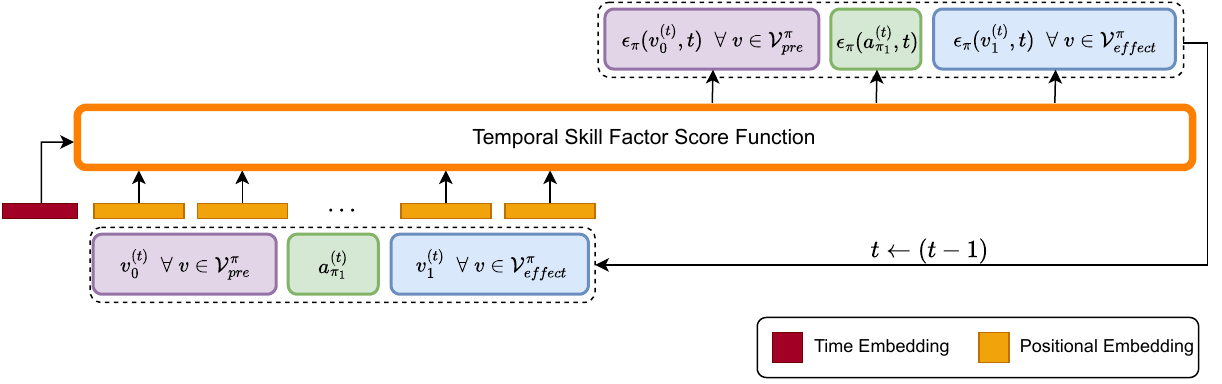}
    \caption{Transformer-based skill diffusion model. We use the noisy pre-condition, action and effect node value distribution at diffusion step t to obtain the corresponding $\epsilon$ during sampling.}
    \label{fig:model_architecture}
\end{figure}



\begin{algorithm*}[h]
    \caption{Training skill score functions for a particular skill $\pi$}
    \SetAlgoLined
    \label{algo:training}
    $\text{ \textbf{Inputs}: }$\\
    $\text{ Pre-condition, skill parameter and Effect nodes } (\mathcal{V}^{\pi}_{pre}, a_\pi, \mathcal{V}^{\pi}_{effect})$ \\
    $\text{ Dataset of transitions } \mathcal{D}$ \\
    $\text{ Parameterized skill score function } \epsilon_\phi$ \\
    $\text{ Noise schedule } \sigma$ \\
    $\text{ DSM loss weight schedule } \lambda$ \\
    \vspace{1em}

    \While{\text{ not converged }}{
    $\text{ Sample batch from dataset } \mathbf{x}^{(0)} \sim \mathcal{D} $\\
    $\text{ Sample forward diffusion timestep } t \sim [0, 1] $\\
    $\text{ Sample Gaussian noise } \epsilon \sim \mathcal{N}(0, \mathbf{I}) $\\
    $\text{ Calculate noise coefficient } \sigma_t$\\
    $\text{ Calculate noisy data } \mathbf{x}^{(t)} = \mathbf{x}^{(0)} + \sigma_t\epsilon$\\
    }

    \vspace{1em}
    $\text{ Optimize parameters } \phi \text{ using: }$\\
    $\;\;\;\;\;\;\;\;\;\;\;\;\nabla_\phi\mathbb{E}_{t,\epsilon, \textbf{x}^{(0)}}[ \lambda(t) \| \epsilon -  \epsilon_\phi(\textbf{x}^{(t)}, t) \|^2 ]  $ \\
    \vspace{1em}
    $\text{Return } \epsilon_\pi \equiv \text{ (Optimized) }\epsilon_\phi $
\end{algorithm*}

\textbf{Hyperparameters and computation.}  We consider the hyperparameters as shown in~\autoref{tab:hparams} for building our score-network.

\begin{table}[h!]
    \centering
    \caption{Hyperparameters for Score-Network with Transformer Backbone}
    \begin{tabular}{|c|c|}
    \hline
       Hyper-parameter  &  Value \\
    \hline
       Hidden Dimension  & 128 \\
       Number of Blocks & 2 \\
       Number of Heads & 2 \\
       MLP Ratio &  2\\
       Dropout Probability & 0.1 \\
       Number of Input Channels & Varies (3-11) \\
       Number of Output Channels & Varies (3-11) \\
       \hline
    \end{tabular}
    \label{tab:hparams}
\end{table}

For the reverse sampling steps while inference, we find the best performance using 50 steps and all results have been reported accordingly. Considering skill-object score functions with varying input nodes leads to a loss of parallel batched inference~(advantage of vectorized states) and hence, an increase in computation time as compared to chaining with vectorized states. On an NVIDIA $\text{RTX}^{\text{TM}}$ A6000 GPU, it takes 2.6 secs for the smallest horizon task \textit{Pour Cup} and 6 secs for the longest horizon task \textit{Hammer Nail} to give 10 candidate node variable values. These candidates are sorted based on their extent of goal-condition satisfaction and the top 5 are selected to calculate the success performance.

\newpage
\section{More Details on Evaluation Tasks}
\label{app:evaluation-tasks}

\subsection{Hammer Nail}

\begin{figure}[h]
    \centering
    \includegraphics[width=\linewidth]{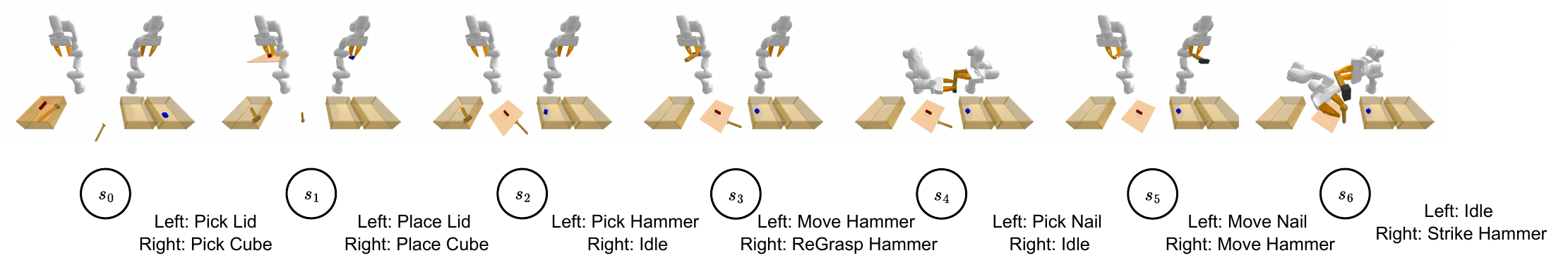}
    \caption{\textbf{Hammer Nail.} The illustration shows the \textit{Hammer Nail} task. A successful solution to this task must complete a successful handover and coordinate to align the hammer and the nail to conduct a successful strike. }
    \label{fig:hammer-sim-chain}
\end{figure}

\textbf{Task Description:} Given a scene with three boxes, a hammer in placed in one of the box covered by a lid as shown in~\autoref{fig:hammer-sim-chain}. There is a nail on the table. Only left arm can reach the lid, hammer and the nail. The task objective is to strike the nail by the hammer within a provided region. There is a cube in one of the boxes, picking and placing it are task-irrelevant distractions.

\textbf{What it takes to solve?} From a superficial symbolic analysis, the task can be completed if the left arm can handover the hammer to the right arm, left arm can pick up the nail to take it to the admissible region and the right arm can strike the nail by the hammer. However, the following challenges exist:
\begin{enumerate}
    \item Hammer must be picked up and moved at a location such that the right arm can re-grasp it for a successful handover.
    \item The handover must allow the right arm to satisfy the pre-condition of strike i.e. the right arm must grasp the hammer away from the head, hence the left arm must reason and pick it up by grasping close to head.
    \item The re-grasp pose will affect the region where the hammer head can be reached. The left arm must reason about the hammer head's reachability to move the nail such that the hammer and nail can be aligned.
\end{enumerate}

\textbf{Why is this challenging?} All the above reasonings are interdependent and the effect of the initial pick pose can be seen at multiple stages of the task. This makes the task challenging as the plans fails:
\begin{enumerate}
    \item if the initial pick pose fails to reason about handover requirements.
    \item if the nail move target pose fails to satisfy the reachability of the hammer-head, which actually depends on the handover.
\end{enumerate}

\begin{figure}[h]
    \centering
    \includegraphics[width=0.9\linewidth]{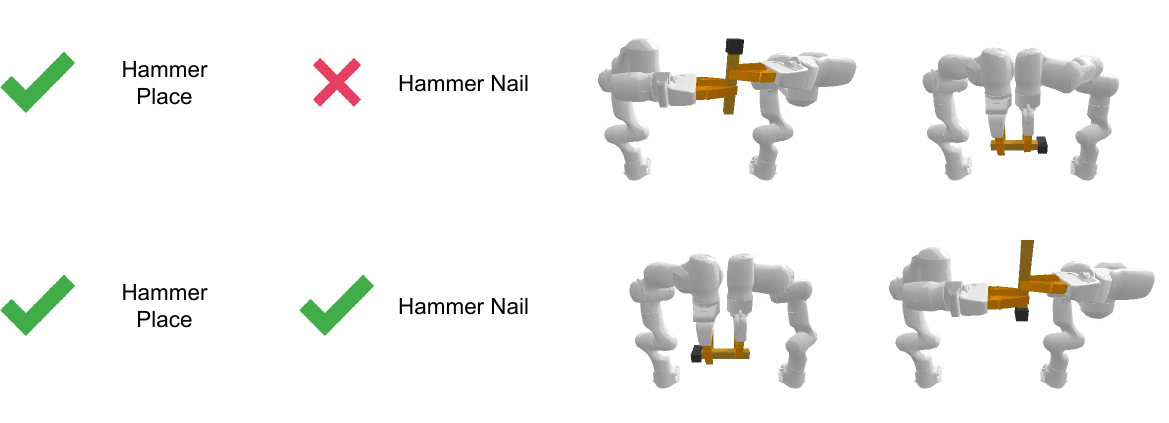}
    \caption{\textbf{Handover variations.} The hammer handover can be done in multiple ways, four of which are shown above. While placement of the hammer in the box for \textit{Hammer Place} task can be done by re-grasping the hammer anywhere, for hammer strike in \textit{Hammer Nail}, the hammer is encouraged to be regrasped near the tail of the handle. }
    \label{fig:enter-label}
    \vspace{-0.3cm}
\end{figure}

\textbf{Failure cases:} The failures in the proposed method occur in the following situations:
\begin{enumerate}
    \item \textit{Method failure:} when it predicts in-feasible poses (where IK cannot be computed) or which does not satisfy the pre-condition of the next skill.
    \item \textit{Trajectory planning failure:} If IK can be computed for current and target poses but no collision-free trajectory can be computed (via pybullet-planning~cite{}). This is expected as GFC only solves for high-level skill transitions.
    \item \textit{Simulation failure:} While executing \texttt{Pick} skill, sometimes the contact vectors are noisy and hence leads to pick-up failures.
\end{enumerate} 

\subsection{Bimanual Pot Reorientation}

\begin{figure}[h]
    \centering
    \includegraphics[width=0.5\linewidth]{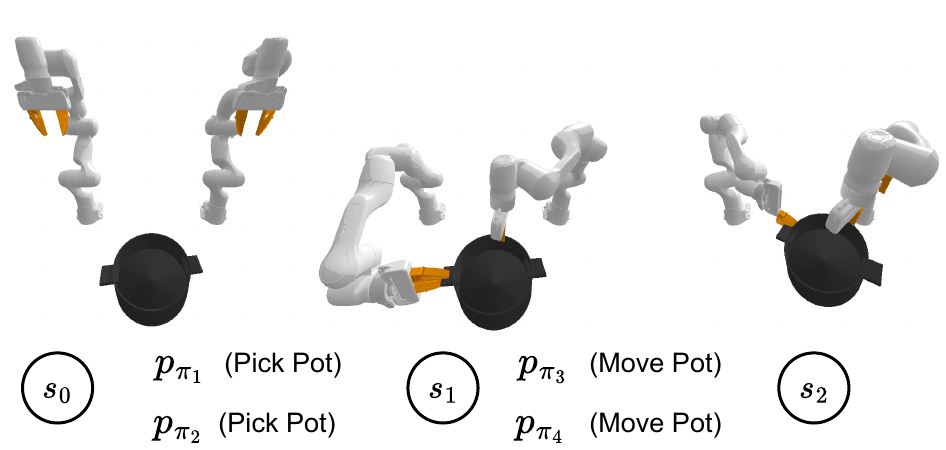}
    \caption{\textbf{Bimanual Pot Reorientation.} The task is to coordinate planning strategies to grasp a pot using two manipulators and rotate it to a target reorientation angle. The task must be done with only single-manipulator data.}
    \label{fig:pot-sim-chain}
    \vspace{-0.2cm}
\end{figure}

\textbf{Task Description:} Given a pot on a table, the task is to reorient the pot to some target orientation angle~(along z-axis) using two manipulators as shown in~\autoref{fig:pot-sim-chain}. It is worth noting that we have \texttt{Pick} and \texttt{Move} skills for individual manipulators such that we know where the pot can be grasped and the reachable workspace of the manipulator.

\textbf{What it takes to solve?} This particular task can be completed if:
\begin{enumerate}
    \item we find pick poses for both the manipulators.
    \item we find feasible move poses in the workspace that satisfies the target orientation.
    \item we ensure that the relative transform between two gripper poses while picking and in the predicted move target poses is the same, because the grasp poses relative to the pot cannot change while moving.
\end{enumerate}

\textbf{Why is this challenging?} The task is challenging because the algorithm must decide the initial pick pose by considering sequential and parallel dependencies:
\begin{enumerate}
    \item the same pick pose relative to the pot must exist for the target reorientation angle
    \item the move pose for both manipulators must satisfy both the workspace reachability for individual manipulators and also have the same fixed transform as the pick poses.
\end{enumerate}

\textbf{Failure cases:} The failures in the proposed method occur in the following situations:
\begin{enumerate}
    \item \textit{Method failure:} when it predicts in-feasible poses (where IK cannot be computed) or which does not satisfy the fixed transform condition.
    \item \textit{Trajectory planning failure:} If IK can be computed for current and target poses but no collision-free trajectory can be computed (via pybullet-planning~cite{}). This is expected as GFC only solves for high-level skill transitions.
    \item \textit{Simulation failure:} While executing \texttt{Pick} skill, sometimes the contact vectors are noisy and hence lead to pick-up failures.
\end{enumerate}

\newpage
\section{Extending Hammer Nail task to longer horizons}
\label{app:hammer-nail-extensions}

\begin{figure}[h]
    \centering
    \includegraphics[width=\linewidth]{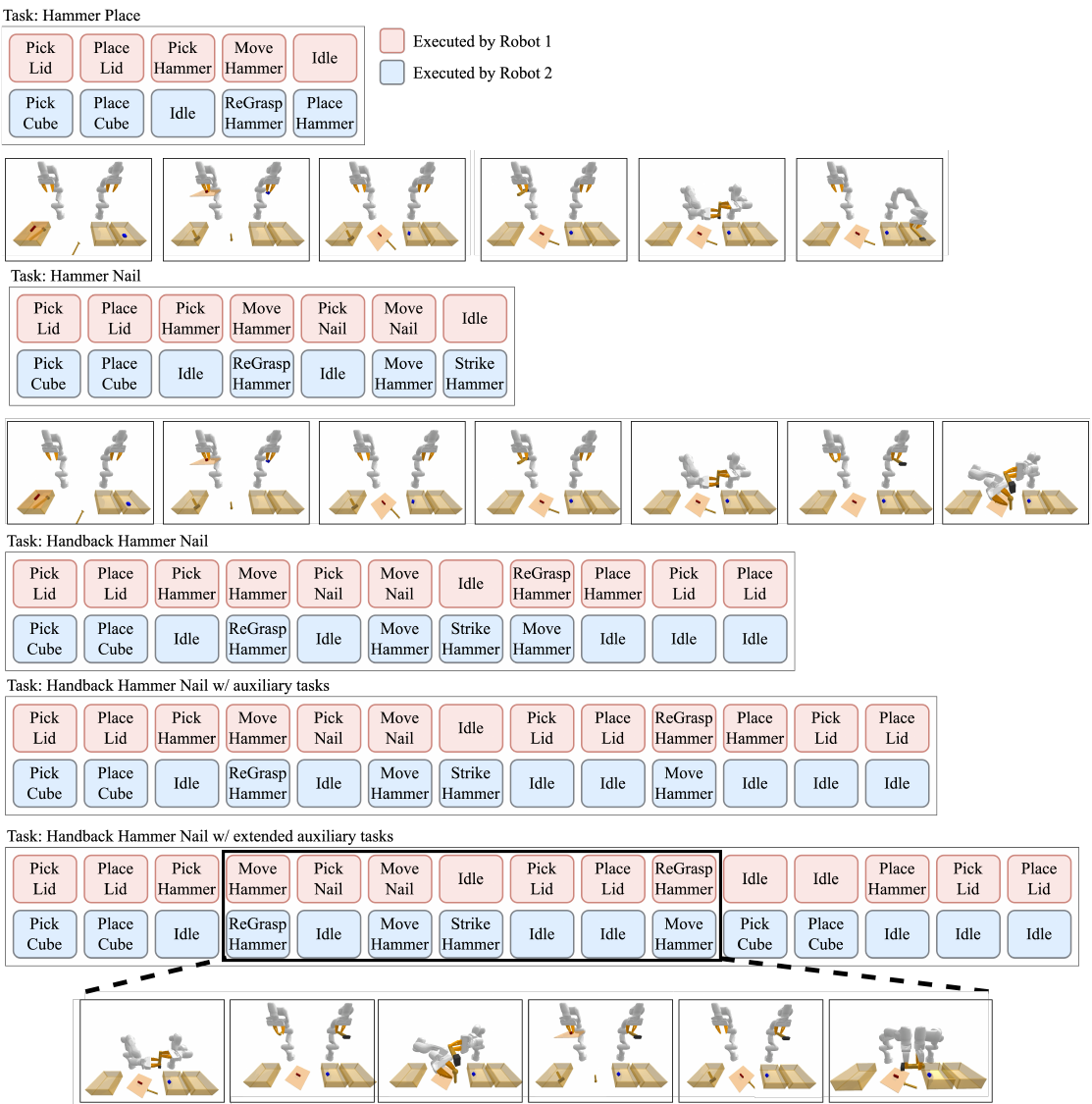}    \caption{\textbf{Extension of Hammer Nail task.} We have added three new extensions to the \textit{Hammer Nail} task. All of the new tasks focus on handling a second handover. The nature of the first handover adds further constraints into possible ways to perform the second handover. Further, we add task-irrelevant skills in between the plan skeleton to evaluate the robustness of GFC and the spatial-temporal factor graph plan representation.}
    \label{fig:hammer-variants}
\end{figure}

In order to evaluate the extensive long-horizon planning capabilities of our proposed algorithm, we have further extended the Hammer Nail task to longer horizons as shown in~\autoref{fig:hammer-variants}. The extended tasks particularly emphasize adding a second handover such that the hammer is handed back to the left arm after a successful hammer strike.

\begin{table}[t]
    \centering
    \caption{Failure breakdown and task success analysis of hammer nail task and its extensions with two handovers (based on 100 trials)}
    \label{tab:overview_breakdown}
    \begin{tabularx}{\textwidth} { 
    |c 
    |c 
      | >{\centering\arraybackslash}X 
      | >{\centering\arraybackslash}X 
      | >{\centering\arraybackslash}X 
      | >{\centering\arraybackslash}X 
      | >{\centering\arraybackslash}X |}
     \hline
     Task & Task Horizon & Type 1 failure & Type 2 failure & Type 3 failure & Task Success \\ \hline
     Hammer Nail & 11 & 42 & 14 & 10 & \textbf{34} \\ \hline
     Extended Hammer Nail v1 & 16 & 43 & 28 & 5 & \textbf{24} \\ \hline
     Extended Hammer Nail v2 & 18 & 44 & 21 & 10 & \textbf{25} \\ \hline
     Extended Hammer Nail v3 & 20 & 41 & 25 & 13 & \textbf{21} \\ \hline
    \end{tabularx}
\end{table}

We classify the failure cases as:
\begin{itemize}
    \item Type 1: Method failure i.e. when the proposed algorithm fails to find suitable target parameters.
    \item Type 2: Trajectory planning failure i.e. no collision-free trajectory can be computed between two suitable poses.
    \item Type 3: Simulation failure i.e. when simulator fails to detect suitable contacts.
\end{itemize}

Now, we show the failure breakdown and task success for all the considered \textit{Hammer Nail} task and their extensions in~\autoref{tab:overview_breakdown}. While we see a drop in success rates by adding a second handover to the vanilla \textit{Hammer Nail} task, GFC proved to be robust for all other task-irrelevant skills in the chain. The task success of all ``two handover" variants is similar even with an increasing task horizon.




\newpage
\section{Justifying success rates with breakdowns}
\label{sec:task-breakdowns}

We elaborate on the failure and success breakdown for the vanilla \textit{Hammer Nail} task in~\autoref{tab:hammer_nail_breakdown}. Revisiting the failure categories, we classify the failure cases as:
\begin{itemize}
    \item Type 1: Method failure i.e. when the proposed algorithm fails to find suitable target parameters.
    \item Type 2: Trajectory planning failure i.e. no collision-free trajectory can be computed between two suitable poses.
    \item Type 3: Simulation failure i.e. when simulator fails to detect suitable contacts.
\end{itemize}

\begin{table}[h]
    \centering
    \caption{Failure breakdown and task success analysis per skill-step of hammer nail task (based on 100 trials)}
    \label{tab:hammer_nail_breakdown}
    \begin{tabularx}{\textwidth} { 
    |c 
    |c 
      | >{\centering\arraybackslash}X 
      | >{\centering\arraybackslash}X 
      | >{\centering\arraybackslash}X 
      | >{\centering\arraybackslash}X 
      | >{\centering\arraybackslash}X |}
     \hline
     Skill.No. & Skills & Type 1 failure & Type 2 failure & Type 3 failure & Accu. Success \\ \hline
     1 & Pick Lid & 5 & 0 & 0 & \textbf{95} \\ \hline
     2 & Place Lid & 0 & 0 & 0 & \textbf{95} \\ \hline
     3 & Pick Cube & 0 & 0 & 0 & \textbf{95} \\ \hline
     4 & Place Cube & 6 & 0 & 0 & \textbf{89} \\ \hline
     5 & Pick Hammer & 3 & 0 & 2 & \textbf{84} \\ \hline
     6-7 & Move Hammer - Regrasp Hammer & 8 & 6 & 0 & \textbf{70} \\ \hline
     8 & Pick Nail & 4 & 0 & 8 & \textbf{58} \\ \hline
     9-10 & Move Nail - Move Hammer & 11 & 8 & 0 & \textbf{39} \\ \hline
     11 & Hammer Strike & 5 & 0 & 0 & \textbf{34}\\\hline
    \end{tabularx}
\end{table}

We also elaborate on the failure and success breakdown for the bimanual reorientation task in~\autoref{tab:pot_breakdown}. It is worth to be noted that the skills are executed in parallel and the serialized representation of the skill sequence is shown only as a part of the analysis.

\begin{table}[h]
    \centering
    \caption{Failure breakdown and task success analysis per skill step of bimanual pot reorientation (based on 100 trials)}
    \label{tab:pot_breakdown}
    \begin{tabularx}{\textwidth} { 
    |c 
    |c 
      | >{\centering\arraybackslash}X 
      | >{\centering\arraybackslash}X 
      | >{\centering\arraybackslash}X 
      | >{\centering\arraybackslash}X 
      | >{\centering\arraybackslash}X |}
     \hline
     Skill.No. & Skills & Type 1 failure & Type 2 failure & Type 3 failure & Accu. Success \\ \hline
     1 & Grasp Pot Left & 13 & 0 & 4 & \textbf{83} \\ \hline
     2 & Grasp Pot Right & 12 & 0 & 3 & \textbf{68} \\ \hline
     3-4 & Move Pot Left - Move Pot Right & 13 & 12 & 0 & \textbf{53} \\ \hline
    \end{tabularx}
\end{table}



We further continue the analysis for all the two handover extensions of the \textit{Hammer Nail} task, namely for \textit{Extended Hammer Nail v1} in~\autoref{tab:hammer_nail_v1_breakdown}, for \textit{Extended Hammer Nail v2} in~\autoref{tab:hammer_nail_v2_breakdown}, and for \textit{Extended Hammer Nail v3} in~\autoref{tab:hammer_nail_v3_breakdown}. We primarily note the accumulative success at the first handover, coordination for the hammer \texttt{Strike}, and the second handover. With an increasing task horizon, the proposed approach is invariant to task-irrelevant distractions and maintains similar success.

\begin{table}[h]
    \centering
    \caption{Failure breakdown and task success analysis per skill-step of hammer nail task extension v1 with two handovers (based on 100 trials)}
    \label{tab:hammer_nail_v1_breakdown}
    \begin{tabularx}{\textwidth} { 
    |c 
    |c 
      | >{\centering\arraybackslash}X 
      | >{\centering\arraybackslash}X 
      | >{\centering\arraybackslash}X 
      | >{\centering\arraybackslash}X 
      | >{\centering\arraybackslash}X |}
     \hline
     Skill.No. & Skills & Type 1 failure & Type 2 failure & Type 3 failure & Accu. Success \\ \hline
     1 & Pick Lid & 4 & 0 & 0 & \textbf{96} \\ \hline
     2 & Place Lid & 0 & 0 & 0 & \textbf{96} \\ \hline
     3 & Pick Cube & 0 & 0 & 0 & \textbf{96} \\ \hline
     4 & Place Cube & 5 & 0 & 0 & \textbf{91} \\ \hline
     5 & Pick Hammer & 4 & 0 & 2 & \textbf{85} \\ \hline
     6-7 & Move Hammer - Regrasp Hammer & 11 & 13 & 0 & \textbf{61} \\ \hline
     8 & Pick Nail & 3 & 0 & 3 & \textbf{55} \\ \hline
     9-10 & Move Nail - Move Hammer & 7 & 9 & 0 & \textbf{39} \\ \hline
     11 & Hammer Strike & 3 & 0 & 0 & \textbf{36}\\\hline
     12-13 & Move Hammer - Regrasp Hammer & 4 & 6 & 0 & \textbf{26}\\\hline
     14 & Place Hammer & 0 & 0 & 0 & \textbf{26} \\
    \hline
    15 & Pick Lid & 2 & 0 & 0 & \textbf{24} \\
    \hline
    16 & Place Lid & 0 & 0 & 0 & \textbf{24} \\
    \hline
    \end{tabularx}
\end{table}

\begin{table}[h]
    \centering
    \caption{Failure breakdown and task success analysis per skill-step of hammer nail task extension v2 with two handovers and some task-irrelevant skills (based on 100 trials)}
    \label{tab:hammer_nail_v2_breakdown}
    \begin{tabularx}{\textwidth} { 
    |c 
    |c 
      | >{\centering\arraybackslash}X 
      | >{\centering\arraybackslash}X 
      | >{\centering\arraybackslash}X 
      | >{\centering\arraybackslash}X 
      | >{\centering\arraybackslash}X |}
     \hline
     Skill No. & Skills & Type 1 failure & Type 2 failure & Type 3 failure & Accu. Success \\ \hline
     1 & Pick Lid & 4 & 0 & 0 & \textbf{96} \\ \hline
     2 & Place Lid & 0 & 0 & 0 & \textbf{96} \\ \hline
     3 & Pick cube &  0 & 0 & 0 & \textbf{96}\\ \hline
     4 & Place Cube & 4 & 0 & 0 & \textbf{92} \\ \hline
     5 & Pick Hammer & 5 & 0 & 2 & \textbf{85} \\ \hline
     6-7 & Move Hammer - Regrasp Hammer & 12 & 14 & 0 & \textbf{59} \\ \hline
     8 & Pick Nail & 2 & 0 & 1 & \textbf{56} \\ \hline
     9-10 & Move Nail - Move Hammer & 4 & 0 & 7 & \textbf{45} \\ \hline
     11 & Hammer Strike & 1 & 0 & 0 & \textbf{44}\\\hline
     12 & Pick Lid & 3 & 0 & 0 & \textbf{41} \\ \hline
     13 & Place Lid & 0 & 0 & 0 &  \textbf{41} \\ \hline
     14-15 & Move Hammer - Regrasp Hammer & 6 & 7 & 0 & \textbf{28} \\\hline
     16 & Place Hammer & 0 & 0 & 0 &  \textbf{28} \\ \hline
     17 & Pick Lid & 3 & 0 & 0 &  \textbf{25} \\ \hline
     18 & Place Lid & 0 & 0 & 0 &  \textbf{25} \\ 
    \hline
    \end{tabularx}
\end{table}

\begin{table}[h]
    \centering
    \caption{Failure breakdown and task success analysis per skill-step of hammer nail task extension v3 with two handovers and many task-irrelevant skills (based on 100 trials)}
    \label{tab:hammer_nail_v3_breakdown}
    \begin{tabularx}{\textwidth} { 
    |c 
    |c 
      | >{\centering\arraybackslash}X 
      | >{\centering\arraybackslash}X 
      | >{\centering\arraybackslash}X 
      | >{\centering\arraybackslash}X 
      | >{\centering\arraybackslash}X |}
     \hline
     Skill No. & Skills & Type 1 failure & Type 2 failure & Type 3 failure & Accu. Success \\ \hline
     1 & Pick Lid & 5 & 0 & 0 & \textbf{95} \\ \hline
     2 & Place Lid & 0 & 0 & 0 & \textbf{95} \\ \hline
     3 & Pick cube & 0 & 0 & 2 & \textbf{93} \\ \hline
     4 & Place Cube & 4 & 0 & 0 & \textbf{89} \\ \hline
     5 & Pick Hammer & 3 & 0 & 2 & \textbf{84} \\ \hline
     6-7 & Move Hammer - Regrasp Hammer & 4 & 8 & 0 & \textbf{72}\\ \hline
     8 & Pick Nail & 3 & 0 & 6 & \textbf{63} \\ \hline
     9-10 & Move Nail - Move Hammer & 7 & 9 & 0 & \textbf{47} \\ \hline
     11 & Hammer Strike & 5 & 0 & 0 & \textbf{42}\\\hline
     12 & Pick Lid & 1 & 0 & 2 & \textbf{39} \\ \hline
     13 & Place Lid & 0 & 0 & 0 & \textbf{39} \\ \hline
     14-15 & Move Hammer - Regrasp Hammer & 5 & 8 & 0 & \textbf{26}\\\hline
     16 & Pick cube & 0 & 0 & 0 & \textbf{26} \\ \hline
     17 & Place Cube &  1 & 0 & 0 & \textbf{25}\\ \hline
     18 & Place Hammer & 3 & 0 & 0 & \textbf{22} \\ \hline
     19 & Pick Lid &  0 & 0 & 1 & \textbf{21}\\ \hline
     20 & Place Lid & 0 & 0 & 0 & \textbf{21} \\
    \hline
    \end{tabularx}
\end{table}

\end{document}